\documentclass{article}

\usepackage{microtype}
\usepackage{graphicx}
\usepackage{subcaption}
\usepackage{booktabs} % for professional tables

\usepackage{hyperref}

\usepackage[accepted]{icml2026}

\usepackage{amsmath}
\usepackage{amssymb}
\usepackage{mathtools}
\usepackage{amsthm}
\usepackage{array}
\usepackage{enumitem}
\usepackage{bm}
\usepackage{url} 
\usepackage{algorithm}
\usepackage{algorithmic}
\usepackage{multirow}

\usepackage[capitalize,noabbrev]{cleveref}

\theoremstyle{plain}
\newtheorem{theorem}{Theorem}[section]
\newtheorem{proposition}[theorem]{Proposition}

\newtheorem{corollary}[theorem]{Corollary}
\theoremstyle{definition}

\theoremstyle{remark}
\newtheorem{remark}[theorem]{Remark}

\icmltitlerunning{Beyond the Proxy: Trajectory-Distilled Guidance for Offline GFlowNet Training}

\begin{document}

\twocolumn[
  \icmltitle{Beyond the Proxy: Trajectory-Distilled Guidance for Offline GFlowNet Training}

  % It is OKAY to include author information, even for blind submissions: the
  % style file will automatically remove it for you unless you've provided
  % the [accepted] option to the icml2026 package.

  % List of affiliations: The first argument should be a (short) identifier you
  % will use later to specify author affiliations Academic affiliations
  % should list Department, University, City, Region, Country Industry
  % affiliations should list Company, City, Region, Country

  \begin{icmlauthorlist}
    \icmlauthor{Ruishuo Chen}{iiis}
    \icmlauthor{Xun Wang}{iiis}
    \icmlauthor{Rui Hu}{iiis}
    \icmlauthor{Zhuoran Li}{iiis}
    \icmlauthor{Longbo Huang}{iiis}
  \end{icmlauthorlist}

  \icmlaffiliation{iiis}{Institute for Interdisciplinary Information Sciences, Tsinghua University, Beijing, China}

  \icmlcorrespondingauthor{Longbo Huang}{longbohuang@tsinghua.edu.cn}

  % You may provide any keywords that you find helpful for describing your
  % paper; these are used to populate the "keywords" metadata in the PDF but
  % will not be shown in the document
  \icmlkeywords{GFlowNets,proxy-free guidance}

  \vskip 0.3in
]

% this must go after the closing bracket ] following \twocolumn[ ...

% This command actually creates the footnote in the first column listing the
% affiliations and the copyright notice. The command takes one argument, which
% is text to display at the start of the footnote. The \icmlEqualContribution
% command is standard text for equal contribution. Remove it (just {}) if you
% do not need this facility.

\printAffiliationsAndNotice{}

\begin{abstract}
Generative Flow Networks (GFlowNets) excel at sampling diverse, high-reward objects. In many practical applications where active reward queries are infeasible, these models must be trained using static offline datasets. Prevailing training methods typically rely on a proxy model to provide reward feedback for online sampled trajectories. However, constructing a reliable proxy is often challenging due to data scarcity or high evaluation costs. While existing proxy-free approaches attempt to address this, they often impose coarse constraints that limit the model's ability to explore effectively. To overcome these limitations, we propose \textbf{Trajectory-Distilled GFlowNet (TD-GFN)}, a novel proxy-free training framework. TD-GFN utilizes inverse reinforcement learning (IRL) to extract dense, transition-level edge rewards from offline trajectories, providing rich structural guidance for efficient exploration. Crucially, to ensure robustness, these rewards guide the policy indirectly through DAG pruning and prioritized backward sampling. This design ensures that gradient updates rely exclusively on ground-truth terminal rewards from the dataset, thereby preventing error propagation. Empirical results demonstrate that TD-GFN significantly outperforms a broad range of existing baselines in both convergence speed and sample quality, establishing a more robust and efficient paradigm for offline GFlowNet training.
\end{abstract}

\section{Introduction}

Generative Flow Networks (GFlowNets or GFNs)~\citep{bengio2021flow, bengio2023gflownet} are a powerful class of generative models designed to sample compositional objects in proportion to their rewards. They have demonstrated significant success across diverse domains, including molecule discovery~\citep{bengio2021flow}, biological sequence design~\citep{jain2022biological}, combinatorial optimization~\citep{zhang2025adversarial}, and generative model fine-tuning~\citep{hu2024amortizing, liu2025efficient}. Despite this success, active environment interaction for reward feedback is often impractical in practical applications due to expensive experiments or the need for human evaluation. This constraint necessitates training GFlowNets from pre-collected offline datasets.

The standard paradigm involves training a proxy model to approximate reward signals from the dataset, which the GFlowNet then queries to facilitate policy optimization. However, constructing a reliable reward model is resource-intensive, typically requiring large, diverse datasets and significant domain expertise~\citep{ouyang2022training, jain2023gflownets}. Furthermore, reward queries targeting out-of-distribution samples, or states inaccurately estimated relative to the proxy, can propagate errors through the gradients and compromise the learned policy. These challenges are especially acute in domains such as language modeling~\citep{ziegler2019fine, dalrymple2024towards}, recommendation systems~\citep{chen2024opportunities}, and autonomous driving~\citep{xie2024advdiffuser}, where obtaining ground-truth rewards is prohibitively expensive because of the high costs associated with expert annotation or experimental trials.

% \begin{figure}[htbp]
%     \includegraphics[width=\linewidth]{figures/stars.pdf}
%     \caption{Comparison of methods on the Molecule Design task using a $1,500$-trajectory dataset. \textit{Top-$10$ Reward} is the average of the $10$ best samples, and \textit{Convergence} is the number of training trajectories required for convergence. We compare against two proxy-based baselines from \citet{bengio2021flow}: Proxy-GFN (using a proxy learned on this dataset) and Oracle-GFN (using the oracle as a best-case proxy). See Section~\ref{sec:mols} for details.}
%     \label{fig:star}
% \end{figure}

Conversely, execution trajectories in these domains are inherently generated and readily available. While often overlooked in proxy-based methods, this trajectory-level information is a cornerstone of offline reinforcement learning~\citep{kumar2020conservative,kostrikov2021offline} that has yet to be fully exploited for GFlowNets. As recent works expand GFlowNets into a broader spectrum of domains~\citep{liu2023generative, hu2024amortizing, zhu2025flowrlmatchingrewarddistributions} to enable diverse generation, a critical question emerges: how can we effectively leverage these offline trajectories to train GFlowNets when the available terminal rewards are insufficient to construct a reliable proxy model?

\begin{figure*}[!t]
    \centering
    \includegraphics[width=0.85\linewidth]{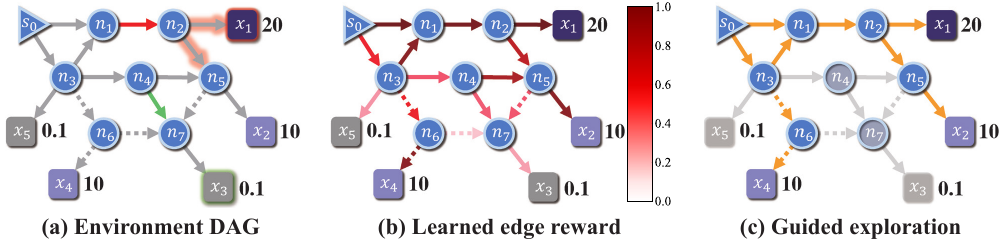}
    \caption{
    Illustration of our motivation. All subfigures depict an environment DAG, where solid edges represent transitions observed in the dataset and dashed edges indicate unobserved transitions. Reward values are annotated next to the terminal states.
    (a) Different edges contribute unequally to the final learned distribution.  
    (b) Edge contributions as inferred by TD-GFN.  
    (c) Selective exploration guided by the learned edge contributions.
}
    \label{fig:key}
\end{figure*}

Recent studies, such as RO-GFlowNets~\citep{wang2023regularized} and COFlowNet~\citep{zhang2025coflownet}, have explored learning directly from offline trajectories to eliminate dependence on proxy models. However, these approaches typically impose coarse constraints to align the policy with the dataset. Such practices can restrict generalization, inhibit effective exploration, and heighten sensitivity to data quality.

Rather than imposing such naive constraints, we introduce \textbf{Trajectory-Distilled GFlowNet (TD-GFN)}, a novel \emph{proxy-free} framework that extracts rich structural information from the environment's directed acyclic graph (DAG) to guide its policy. Our work is based on an insight that different edges in the DAG contribute unequally to learning an effective GFlowNet policy, an observation also shared by prior work~\citep{silva2025when} in different contexts. For example, as shown in Figure~\ref{fig:key}(a), removing edge \((n_1 \rightarrow n_2)\) blocks access to a terminal node with reward \(20\), whereas removing edge \((n_4 \rightarrow n_7)\) only affects a low-reward node (\(0.1\)). By leveraging the theoretical equivalence between GFlowNet training and reinforcement learning~\citep{tiapkin2024generative}, we employ inverse reinforcement learning (IRL) on a rebalanced offline dataset to quantify this varying edge importance. This approach allows us to distill transition-level scores termed edge rewards, as illustrated in Figure~\ref{fig:key}(b). These learned rewards provide dense, intermediate guidance that enables the policy to generalize across the DAG, fundamentally differing from proxy models in both learning objective and application.

To ensure robustness, we utilize these edge rewards in an \emph{indirect} yet effective manner. First, we prune low-utility transitions based on their edge rewards, yielding a more compact and expressive action space that prioritizes high-value trajectories (Figure~\ref{fig:key}(c)). Second, we introduce a prioritized backward sampling procedure guided by both terminal and edge rewards, which strategically allocates the model's attention across terminal states and intermediate pathways during training. Throughout this process, gradient-based updates to our policy rely exclusively on ground-truth terminal rewards from the dataset. This indirect application insulates the policy from potential inaccuracies in the IRL phase, preventing the error propagation that plagues proxy-based methods. More importantly, the dense structural guidance provided during decision-making enables TD-GFN to explore high-reward samples with superior efficiency.

Extensive experiments confirm these advantages, demonstrating that our proposed TD-GFN trains with superior efficiency and
reliability, significantly outperforming a wide range of baselines in convergence speed and final
sample quality. These results establish TD-GFN as a more robust, stable, and efficient paradigm for
offline GFlowNet training, marking a significant step forward from conventional methods that unlocks
the potential for deploying GFlowNets in a broader spectrum of complex domains and scenarios.

In summary, our key contributions are as follows:
\begin{itemize}
    \item \textbf{A Novel Proxy-Free Paradigm:} We introduce TD-GFN, a paradigm for training GFlowNets from offline trajectory data that eliminates dependence on proxy models and avoids out-of-distribution reward queries. By employing IRL to distill transition-level edge rewards from a rebalanced dataset, we provide structured guidance that to guide the policy’s training and generalization across the environment.

    \item \textbf{Robust Guidance Mechanism:} We propose two mechanisms for indirect policy guidance using learned edge rewards. Through DAG pruning and prioritized backward sampling, TD-GFN facilitates efficient training under dense guidance while ensuring that gradient updates depend exclusively on ground-truth rewards, thereby preventing error propagation.

    \item \textbf{State-of-the-Art Performance and Efficiency:} Through extensive experiments, we demonstrate that TD-GFN establishes a new state-of-the-art for offline GFlowNet training. Our method significantly outperforms diverse baselines by achieving a superior fit to the target distribution, faster convergence, and higher sample quality across multiple benchmarks.
\end{itemize}

\noindent\textbf{Code Availability.}\, Source code and scripts to reproduce all experiments are available at \url{https://github.com/Chenruishuo/TD-GFN}.

\section{Preliminaries}
In this section, we describe Generative Flow Networks (GFlowNets or GFNs) and the maximum causal entropy inverse reinforcement learning (IRL) framework, which help clarify our proposed method. Additional related works are provided in Appendix~\ref{apd:related} for completeness.
\subsection{GFlowNets}
GFlowNets~\citep{bengio2021flow,bengio2023gflownet} are formulated on an environment structured as a directed acyclic graph (DAG), denoted by \( G = (V, E) \), where \( V \) and \( E \) represent the sets of nodes and directed edges, respectively. A node \( s' \) is defined as a child of \( s \) if there exists a directed edge \( (s \rightarrow s') \in E \); in this case, \( s \) is considered a parent of \( s' \). The graph comprises a unique root node \( s_0 \in V \) with no incoming edges and a subset of terminal nodes \( \mathcal{X} \subseteq V \) with no outgoing edges.

A positive reward function \( R(x) \) is defined over the terminal nodes \( \mathcal{X} \). GFlowNets aim to learn a stochastic forward policy \( \mathcal{P}_F(s'|s) \) such that the probability of reaching any terminal node \( x \) is proportional to its reward \( R(x) \). This objective is achieved by generating sequences of transitions where the agent selects a child node at each step. We assume deterministic environment dynamics where each action uniquely determines the next state by traversing a specific DAG edge, following standard conventions in GFlowNet literature~\citep{bengio2021flow,jain2022biological,bengio2023gflownet}.

Recent work~\citep{tiapkin2024generative} establishes a formal equivalence between GFlowNet training and entropy-regularized RL under specific conditions: a discount factor \( \gamma = 1 \), an entropy regularization coefficient \( \lambda = 1 \), and a modified reward function. This reward is determined by a fixed backward policy \( \mathcal{P}_B(s|s') \) that samples parent nodes in the DAG in reverse from a child node:
\begin{equation}
\widetilde{R}(s, s') =
\begin{cases}
\log \mathcal{P}_B(s|s'), & \text{if } s \notin \mathcal{X} \cup \{s_f\}, \\
\log R(s),                    & \text{if } s \in \mathcal{X}, \\
0,                            & \text{if } s = s_f,
\end{cases}
\label{eq:new_rew}
\end{equation}
where \( s_f \) denotes an additional absorbing state. Consequently, learning the forward policy reduces to solving the following entropy-regularized policy optimization problem:
\begin{equation}
\mathcal{P}_F^* = \mathop{\mathrm{arg\,max}}_{\pi} \left( \lambda \mathcal{H}(\pi) + \mathbb{E}_{\pi} \left[ \sum_{t=0}^\infty \gamma^t \widetilde{R}(s_t, s_{t+1}) \right] \right),
\label{eq:equiva}
\end{equation}
where \( \mathcal{H}(\pi) = \mathbb{E}_\pi \left[ -\log \pi(s'|s) \right] \) denotes the causal entropy of the policy. The expectation \( \mathbb{E}_{\pi} \) is taken with respect to trajectories generated by \( s_0 \sim P_0 \) and \( s_{t+1} \sim \pi(\cdot|s_t) \), where \( P_0 \) is the initial state distribution that corresponds to a Dirac delta distribution in the GFlowNet setting.

In this paper, we address the problem of training a GFlowNet using an offline dataset \( \mathcal{D} = \{ \tau_i = (s_0, s_1, \dots, s_{T_i}, R(s_{T_i})) \}_{i=1}^M \), which consists of trajectories of constructed objects generated by an unknown behavior policy, along with their corresponding ground-truth rewards.

\subsection{Maximum causal entropy IRL}
Maximum causal entropy inverse reinforcement learning~\citep{ziebart2008maximum,ziebart2010modeling} aims to recover a reward function \( r(s, s') \) that both characterizes expert behavior under a policy \( \pi_E \) and maximizes the causal entropy of the learned policy. This objective is formalized as follows:
\begin{equation}
\begin{aligned}
\underset{r}{\text{minimize}} \quad & \Biggl( \max_\pi \left( \lambda \mathcal{H}(\pi) + \mathbb{E}_{\pi} \left[ \sum_{t=0}^\infty \gamma^t r(s_t, s_{t+1}) \right] \right) \\
& - \mathbb{E}_{\pi_E} \left[ \sum_{t=0}^\infty \gamma^t r(s_t, s_{t+1}) \right] \Biggr),
\end{aligned}
\label{eq:IRL}
\end{equation}
where \( \mathcal{H}(\pi) \) denotes the causal entropy of policy \( \pi \), consistent with the definition in Equation~\eqref{eq:equiva}. In our framework, we adapt this objective to infer an edge-level reward function \( R_E: E \rightarrow \mathbb{R} \) that reflects the expert policy's preferences for specific transitions within the graph.

\section{Method}
In this section, we introduce \textbf{Trajectory-Distilled GFlowNet (TD-GFN)}, a novel proxy-free training framework. In contrast to existing methods that restrict the policy to the dataset distribution~\citep{wang2023regularized,zhang2025coflownet}, TD-GFN facilitates a superior balance between exploration and exploitation by strategically allocating attention across the environment DAG.

\begin{figure}[ht]
    \centering    \includegraphics[width=\linewidth]{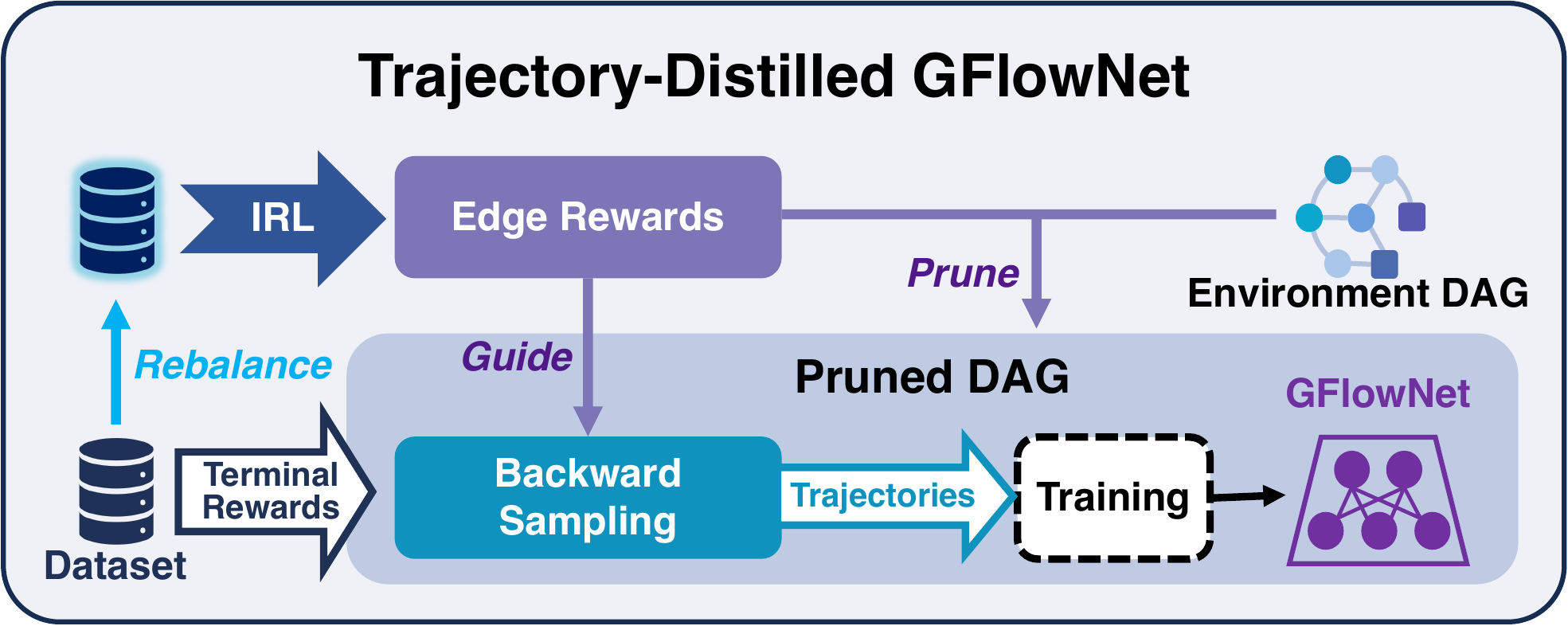}
    \caption{
    An overview of the complete training pipeline of the TD-GFN framework.
    }
    \label{fig:ppl}
\end{figure}

The training pipeline is illustrated in Figure~\ref{fig:ppl}. Initially, TD-GFN extracts an \emph{edge reward function} from a rebalanced dataset using inverse reinforcement learning (IRL). This function captures the relative importance of transitions, providing principled guidance for policy optimization. Using these rewards, TD-GFN prunes the DAG to eliminate low-utility transitions and optimizes the policy through prioritized backward sampling on the resulting subgraph. This design enables effective generalization to unseen regions of the state space while maintaining alignment with high-reward behaviors. The following subsections detail each component, and a complete summary of the procedure is provided in Algorithm~\ref{alg:full} (Appendix~\ref{apd:code}).

\subsection{Edge rewards extraction via IRL}
\label{sec:edge_reward}
Since different edges in the environment DAG contribute unequally to policy learning, we aim to quantify this disparity to enable more targeted and generalizable training. However, measuring the contribution of individual transitions under stochastic GFlowNet policies remains challenging. To address this, we leverage the theoretical equivalence between GFlowNet training and entropy-regularized reinforcement learning (Equation~\eqref{eq:equiva}). Specifically, we apply the maximum causal entropy IRL framework~\citep{ziebart2008maximum,ziebart2010modeling} (Equation~\eqref{eq:IRL}) to extract a fine-grained edge-level reward function from offline trajectories. These learned edge rewards act as structured auxiliary signals that capture the relative importance of transitions, guiding the policy to generalize effectively across the environment DAG.

A primary challenge in applying IRL to offline data is that the underlying behavior policy may not represent expert behavior. To address this, we introduce a rebalancing strategy inspired by~\citet{hong2023harnessing} that biases trajectory sampling toward high-reward terminal states. Specifically, when estimating expectations under the expert policy \( \mathbb{E}_{\pi_E} \), we sample trajectories \( \tau = (s_0, s_1, \dots, s_T) \in \mathcal{D} \) with probability \( P(\tau) \propto R(s_T) \). This approach constructs a \emph{rebalanced dataset} that approximates the visitation distribution of an expert policy. By adjusting edge visitation frequencies to favor high-reward regions, this reweighting scheme effectively creates a pseudo-expert dataset and aligns the learned edge rewards more closely with expert behavior.

Building on the well-known maximum causal entropy IRL framework GAIL~\citep{ho2016generative} which has been theoretically and empirically shown to effectively learn reward functions with strong generalization~\citep{Xu2020ErrorBO,luo2022transferable,luo2024rewardconsistent}, we adopt the following adversarial training objective based on the rebalanced dataset:
\begin{equation}
\begin{aligned}
    \min_{\phi} \max_{\psi}\, &\mathcal{L}(\psi,\phi) = \mathbb{E}_{s\sim \widetilde{\mathcal{D}},s'\sim \pi_\psi} \left[\log D_\phi(s,s')\right] \\
    + &\mathbb{E}_{(s,s')\sim \widetilde{\mathcal{D}}}[\log(1-D_\phi(s,s'))] + \lambda\mathcal{H}(\pi_\psi),
\end{aligned}
\label{eq:obj}
\end{equation}
where \( \widetilde{\mathcal{D}} \) represents the distributions of nodes and edges  (with slight abuse of notation), while \( \pi_\psi \) is a parameterized policy and \( D_\phi: E \rightarrow (0,1) \) is a discriminative classifier. Both components are optimized iteratively.

However, the original GAIL framework induces a strictly non-negative reward function, rendering it incapable of recovering the true underlying reward in our setting, which may include negative values as shown in Equation~\eqref{eq:new_rew}. To address this, we follow~\citet{kostrikov2018discriminator} and extract an unbiased edge-level reward function \( R_E \) from the classifier:
\begin{equation}
    R_E(s,s') = \log D_\phi(s,s') - \log\left(1 - D_\phi(s,s')\right)
    \label{eq:R_E}
\end{equation}

Intuitively, the learned edge reward reflects the preference of the near-expert behavior policy implied by the rebalanced dataset, thereby quantifying the importance of each transition for learning a high-quality GFlowNet policy. This approach differs fundamentally from proxy reward models because it is not designed to predict terminal rewards. Instead, it provides indirect, non-gradient guidance to the policy, as detailed in the following section. The complete procedure for edge reward extraction is summarized in Algorithm~\ref{alg:edge_reward} (Appendix~\ref{apd:code}).

To understand what $R_E$ recovers in the population limit, we analyze Equation~\eqref{eq:IRL} under standard finite-horizon GFlowNet conventions (Appendix~\ref{apd:theory}). Let $\widetilde{R}(x) := w(x) R(x)$ denote the terminal weight of the rebalanced expert. By Proposition~\ref{prop:reweight}, this rebalanced expert shares the backward policy $\mathcal{P}_B^*$ of the original GFlowNet, so the canonical edge reward $r^*(s, s') := \log \mathcal{P}_B^*(s \mid s')$ on $E$ is invariant to rebalancing.

\begin{theorem}[Unique Recovery of the Backward Policy]
\label{thm:body-unique}
Restrict the population IRL problem in Equation~\eqref{eq:IRL} to rewards $r$ on $\bar{E}$ satisfying the terminal-boundary condition $r(x, s_f) = \log \widetilde{R}(x)$ for every $x \in \mathcal{X}$. Among all global minimizers, the unique reward also satisfying the parent-normalization gauge
\[
  \sum_{p \in \mathrm{Pa}(s')} \exp\!\bigl(r(p, s')\bigr) \;=\; 1 \qquad \forall\, s' \in V \setminus \{s_0\}
\]
is the canonical edge reward $r^*$. Since $r^*$ is itself parent-normalized, the parent softmax in Equation~\eqref{eq:P_B} applied to $r^*$ recovers $\mathcal{P}_B^*$ exactly. More generally, if the learned $R_E$ satisfies $\|R_E - r^*\|_\infty \le \varepsilon$, the recovered backward policy $\hat{\mathcal{P}}_B$ from Equation~\eqref{eq:P_B} obeys
\[
  \sup_{s' \in V \setminus \{s_0\}} \mathrm{TV}\!\bigl(\hat{\mathcal{P}}_B(\cdot \mid s'),\, \mathcal{P}_B^*(\cdot \mid s')\bigr) \;\le\; \tfrac{1}{2}\bigl(e^{2\varepsilon} - 1\bigr).
\]
Proofs appear in Appendix~\ref{apd:theory} (Theorem~\ref{thm:unique} and Theorem~\ref{thm:consistency}).
\end{theorem}

Theorem~\ref{thm:body-unique} shows that $R_E$ targets the log conditional incoming-flow score and gives a quantitative recovery guarantee for the induced backward policy. Empirically, this recovery also generalizes beyond the observed transitions. As illustrated in Figure~\ref{fig:key}(b) and Appendix~\ref{apd:visual}, the learned edge rewards demonstrate strong generalization by assigning meaningful values to transitions absent from the dataset. This property encourages the policy to distribute attention throughout the environment DAG instead of remaining restricted to regions near the training data, a limitation common in prior methods~\citep{wang2023regularized,zhang2025coflownet}. This generalization capability is essential for the GFlowNet objective because accurately matching the target distribution requires the policy to navigate and evaluate states not explicitly represented in the offline data. Notably, this global edge-level guidance aligns with the core GFlowNet philosophy of using structured, multi-step decision-making to decompose complex sampling problems instead of attempting to model the distribution directly.

\subsection{Reward-guided pruning and prioritized backward sampling}
Although edge rewards provide meaningful transition-level signals, using them to shape policy gradients directly increases sensitivity to function approximation errors. This risk mirrors the error propagation issues observed with proxy rewards. To improve both robustness and efficiency, we introduce two mechanisms to shape the GFlowNet policy as illustrated in Figure~\ref{fig:ppl}. Specifically, we perform \textit{reward-guided pruning} on the environment DAG and subsequently train the model using trajectories generated through \textit{prioritized backward sampling} on the resulting pruned graph.

Guided by the learned edge rewards, TD-GFN removes low-utility edges from the environment DAG \(G = (V, E)\). We approximate the distribution of edge rewards using a batch \(\mathcal{D}_{R_E} = \{R_E(s_i, \text{selected}(\pi_\psi(\cdot \mid s_i)))\}_{i=1}^{\mathcal{B}}\), where \(\pi_\psi\) is the imitation policy from Algorithm~\ref{alg:edge_reward} serving as an expert surrogate. A transition \((s \rightarrow s')\) is pruned from the graph if it lies in the low-density region of this reward distribution:
\begin{equation}
    R_E(s, s') < \text{mean}(\mathcal{D}_{R_E}) - K \cdot \text{std}(\mathcal{D}_{R_E}),
\label{eq:prune}
\end{equation}
where \(K\) is a pruning threshold hyperparameter and \(\text{mean}(\cdot)\) and \(\text{std}(\cdot)\) denote the empirical mean and standard deviation over the batch \(\mathcal{D}_{R_E}\). This threshold-based strategy relies on the model's ability to distinguish low-reward edges from potentially useful ones. This approach protects the policy from numerical errors in the IRL phase and reduces the risk of performance degradation. We provide a sensitivity analysis for \(K\) in Appendix~\ref{apd:K-sensitivity} and discuss alternative soft guidance mechanisms in Appendix~\ref{apd:discussion}.

To analyze the robustness of this thresholding rule, we work with a path-based criterion. For each terminal $x \in \mathcal{X}$, define the \emph{bottleneck score}
\[
  \beta(x) \;:=\; \max_{P \in \mathcal{P}(x)} \,\min_{e \in P}\, r^*(e),
\]
where $\mathcal{P}(x)$ denotes the set of directed paths from $s_0$ to $x$. The bottleneck score is the largest minimum edge score along any root-to-$x$ path under $r^*$; high $\beta(x)$ means $x$ is reachable through an edge-score margin that absorbs estimation noise.

\begin{theorem}[Robust Preservation and Distortion]
\label{thm:body-pruning}
Let $R_E$ be a learned edge reward with sup-norm error $\|R_E - r^*\|_\infty \le \varepsilon$, and let $\tau \in \mathbb{R}$ be any pruning threshold (Equation~\eqref{eq:prune} instantiates $\tau = \mathrm{mean}(\mathcal{D}_{R_E}) - K \cdot \mathrm{std}(\mathcal{D}_{R_E})$). Let $\mathcal{X}'$ denote the set of terminals that remain reachable from $s_0$ in the pruned graph $G' = (V, \{e \in E : R_E(e) \ge \tau\})$. Then:
\begin{enumerate}[label=(\roman*),leftmargin=*,itemsep=2pt,topsep=2pt]
\item (Survival.) Every $x \in \mathcal{X}$ with $\beta(x) > \tau + \varepsilon$ satisfies $x \in \mathcal{X}'$.
\item (Distortion.) Assume $\mathcal{X}' \neq \varnothing$, and let $p(x) := R(x)/Z$, $p_\tau(x) := R(x) \cdot \mathbf{1}\{x \in \mathcal{X}'\} / (Z - \Delta_\tau)$ with $\Delta_\tau := \sum_{x \notin \mathcal{X}'} R(x)$ and $\delta_\tau := \Delta_\tau / Z$. Then
\[
  \|p - p_\tau\|_{\mathrm{TV}} \;=\; \delta_\tau \;\le\; \mathbb{P}_{x \sim p}\!\bigl[\beta(x) \le \tau + \varepsilon\bigr],
\]
and $p_\tau(x) / p_\tau(y) = R(x)/R(y)$ for every $x, y \in \mathcal{X}'$.
\end{enumerate}
Proofs appear in Appendix~\ref{apd:theory} (Theorems~\ref{thm:robust} and~\ref{thm:distortion}; Corollary~\ref{cor:mode}).
\end{theorem}

For TD-GFN, high-bottleneck modes therefore survive any reasonable choice of $K$, and on the surviving terminals the original reward-proportional target is preserved exactly up to renormalization. Following this criterion, we remove all edges disconnected from the root node $s_0$, yielding a pruned subgraph $G' = (V', E')$ and an updated terminal set $\mathcal{X}'$. Notably, pruning occurs over the entire environment DAG instead of being restricted to observed transitions. This allows the model to retain paths that may lead to high-reward regions even if they are unobserved in the dataset.

This strategy focuses learning on informative regions and improves efficiency by reducing model fitting complexity~\citep{zahavy2018learn,zhang2020automatic}. The resulting subgraph defines a compact yet expressive action space that prioritizes high-reward trajectories. As illustrated in Figure~\ref{fig:key}(c), this approach facilitates generalization beyond offline data while remaining anchored in reliable signals.

To improve training efficiency on the pruned graph, we propose a prioritized backward sampling mechanism that constructs trajectories in reverse from terminal nodes. Terminal states are sampled from the intersection of $\mathcal{X}'$ and the set of dataset objects $\{s_{T_i}\}_{i=1}^M$ with probability proportional to their ground-truth rewards. From each sampled terminal node $x$, we recursively perform backward sampling toward the root $s_0$ using a learned backward policy $\mathcal{P}_B$, defined as:
\begin{equation}
\mathcal{P}_B(s_t|s_{t+1}) = \frac{\exp{\{R_E(s_t, s_{t+1})\}}}{\sum_{(s, s_{t+1}) \in E'} \exp{\{R_E(s, s_{t+1})\}}},
\label{eq:P_B}
\end{equation}
This formulation aligns with the reward shaping principles in Equation~\eqref{eq:new_rew}. By directing policy updates toward regions prioritized by both terminal and edge rewards, this strategy reinforces the GFlowNet inductive bias of allocating sampling effort in proportion to reward, thereby enhancing sample efficiency. Moreover, combined with the pruning method above, this approach helps mitigate the under-exploitation of terminal nodes discussed in \citet{jang2024pessimistic}.

Finally, the policy is optimized on the pruned DAG using the sampled trajectories and their terminal rewards from the offline dataset. Crucially, the gradient-based updates rely exclusively on these recorded ground-truth values. This approach sidesteps the fitting errors inherent in proxy reward models and insulates the gradients from potential inaccuracies in the edge rewards learned during the IRL phase.

By leveraging reward-guided pruning to steer the policy away from low-utility regions and employing prioritized backward sampling to strategically allocate attention, we effectively achieve exploration of unobserved high-utility regions during evaluation. This paradigm is fundamentally distinct from the trial-and-error exploration inherent in proxy-based methods. Notably, the contributions of TD-GFN are orthogonal to the specific training strategies employed after trajectory sampling, as discussed in Appendix~\ref{apd:ortho}.

\section{Experiments}
In this section, we empirically demonstrate the effectiveness of the proposed TD-GFN through extensive experiments. Specifically, we aim to answer the following questions:  
(i) Can TD-GFN effectively sample from the target reward distribution, and generate high-reward and diverse samples?  
(ii) Is TD-GFN training efficient?  
(iii) Does TD-GFN consistently improve performance across different scenarios?
We also conduct ablation studies to evaluate the individual contributions of each component within our method, as presented in Appendix~\ref{apd:ablation}. Comprehensive task descriptions and implementation details of the algorithm are provided in Appendix~\ref{apd:details}. Further experimental results on additional real-world datasets can be found in Appendix~\ref{apd:rec}.

\subsection{Hypergrid}
\label{sec:hypergrid}
We begin with experiments on the Hypergrid task~\citep{bengio2021flow}, a $D$-dimensional grid environment with side length $H$, resulting in a discrete space of size $H^D$. High-reward modes are narrowly concentrated near the $2^D$ corners, making the task particularly challenging for both exploration-based and offline learning algorithms.
Our main experiments are conducted on a \(8^4\) Hypergrid (\(H = 8\), \(D = 4\)) following the setup in \citet{bengio2021flow}. We also provided results on larger Hypergrids in Appendix~\ref{apd:larger}. Additionally, in Appendix~\ref{apd:visual}, we visualize a \(8 \times 8\) Hypergrid instance and illustrate the TD-GFN workflow on this task.

Due to the nascent nature of proxy-free GFlowNet training, we evaluate TD-GFN against a diverse set of baselines. These include the primary existing proxy-free method, COFlowNet~\citep{zhang2025coflownet}, as well as popular offline RL algorithms such as CQL~\citep{kumar2020conservative} and IQL~\citep{kostrikov2021offline}. We also compare against imitation learning methods, specifically Behavior Cloning (BC) and GAIL~\citep{ho2016generative}. Additionally, we include \textit{Dataset-GFN}, which serves as a vanilla proxy-free variant by training a GFlowNet directly on offline trajectories without modification.
To ensure a fair comparison with COFlowNet, which relies on the flow matching (FM) objective~\citep{bengio2021flow}, we adopt the FM objective (Equation~\eqref{eq:fm}) for our experiments. Because TD-GFN's pruning and prioritized backward sampling operate before the objective computation, they are architecturally orthogonal to the GFlowNet objective: Appendix~\ref{apd:objectives} reports comparable gains when TD-GFN is paired with TB, SubTB, and DB, and Appendix~\ref{apd:naive_offline} contrasts these against the same objectives applied naively to offline data. Furthermore, we train GAIL using the rebalanced dataset described in Section~\ref{sec:edge_reward}. This rebalanced approach improves performance relative to the version trained on the original dataset, and we evaluate it using the imitation policy from Algorithm~\ref{alg:edge_reward}.

Each algorithm is evaluated using two standard metrics~\citep{bengio2021flow,pan2022generative,zhang2023distributional,zhang2025coflownet}. The first is \textit{Empirical L1 Error}, which measures the $\mathcal{L}_1$ distance between the empirical distribution \(\pi(x)\) from model samples and the normalized reward distribution \(p(x) = R(x) / Z\), computed as \(\mathbb{E}[|\pi(x) - p(x)|]\). The second is \textit{Modes Found}, defined as the number of distinct reward modes discovered during training.

We construct an \textit{Expert} dataset comprising $1,500$ trajectories using a well-trained GFlowNet policy to simulate data typically collected by domain experts. We adopt GFlowNet-generated datasets to provide a controlled and reproducible environment for simulating behavior policies with different optimality levels. This strategy is standard within the offline reinforcement learning community~\citep{Fu2020D4RLDF,Gulcehre2020RLUB}. As demonstrated in Section~\ref{sec:bioseq}, TD-GFN also delivers compelling results on real-world data. We provide a more detailed discussion of this approach and an analysis of the value of trajectory structure in Appendix~\ref{apd:discussion}.

\begin{figure}[ht]
    \centering    
    \includegraphics[width=\linewidth]{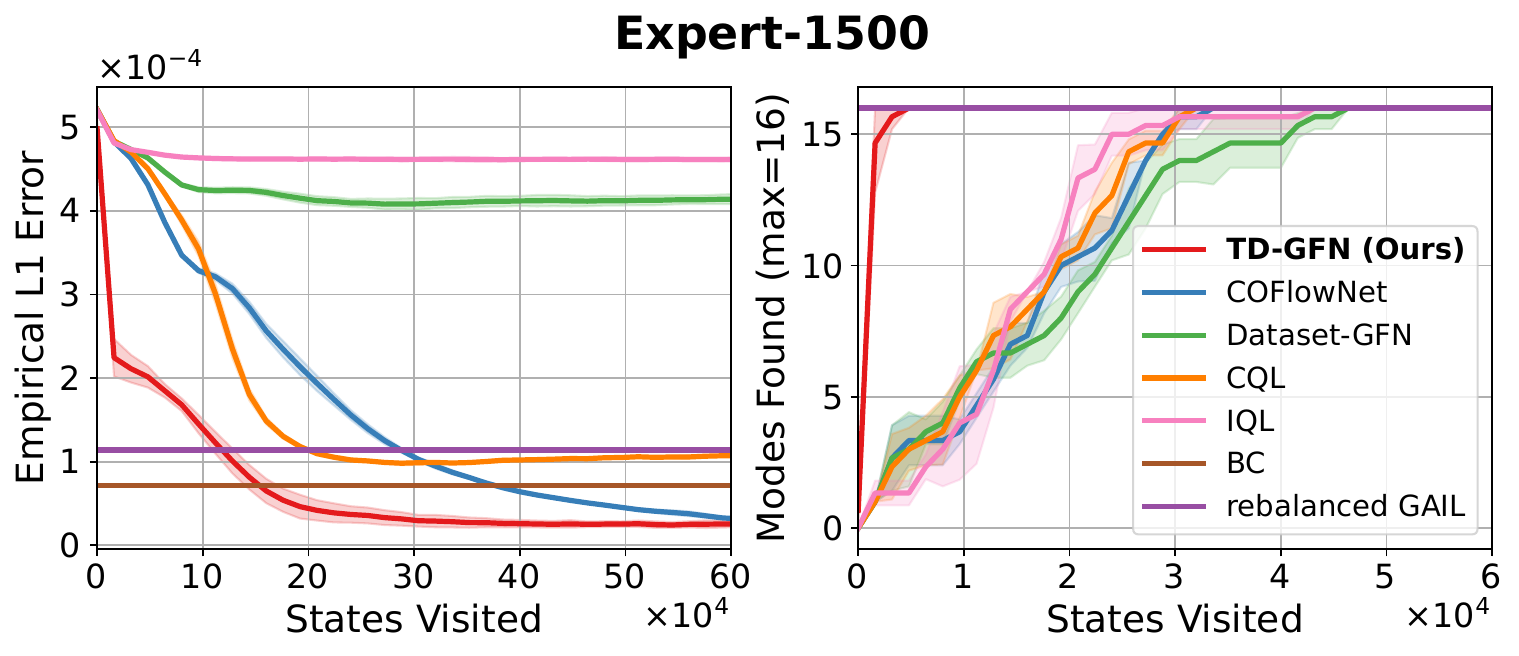}
    \caption{Performance comparison on the $8^4$ Hypergrid task. Results are obtained using $1,500$ trajectories from the \textit{Expert} dataset for policy training. The solid line and shaded region represent the mean and standard deviation, respectively. This convention is maintained for all subsequent figures.}
    \label{fig:grid}
\end{figure}

As illustrated in Figure~\ref{fig:grid}, TD-GFN exhibits superior sample efficiency. It identifies all $16$ modes with fewer than $5,000$ state visits, which represents a \textbf{six-fold} improvement over baselines. Furthermore, it converges to the lowest \textit{Empirical L1 Error} over twice as fast as the most competitive baseline, COFlowNet. We include a comprehensive analysis of the other baselines in Appendix~\ref{apd:baselines}.

\begin{figure}[ht]
    \centering
    \includegraphics[width=\linewidth]{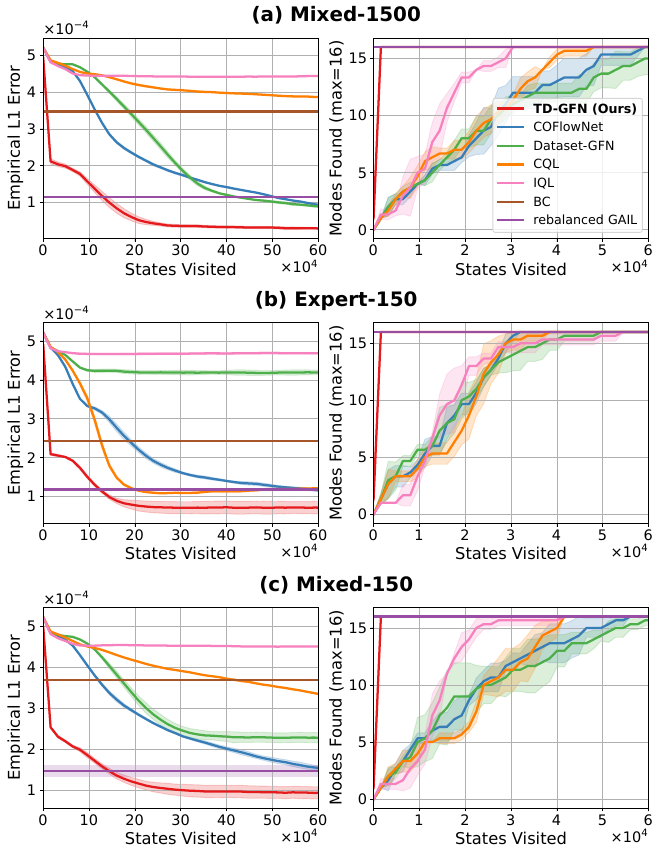}
    \caption{Performance comparison on the $8^4$ Hypergrid task. Labels in the format \textit{Dataset}-\(N\) indicate training conducted with \(N\) trajectories from the specified dataset.}
    \label{fig:grid_noise}
\end{figure}

\textbf{Robustness to Dataset Uncertainty.}\,
We first assess the robustness of TD-GFN under dataset uncertainty by introducing greater variability in the training data. In the \textit{Mixed} setting, we augment the \textit{Expert} dataset with an equal number of trajectories generated by a random policy similar to \citet{hong2023harnessing,cao2024offline}, thereby introducing noise. In another setting, we significantly reduce the number of available trajectories—using only one-tenth of the original dataset—to simulate data scarcity as an additional source of uncertainty~\citep{depeweg2018decomposition}.

As illustrated in Figure~\ref{fig:grid_noise}, TD-GFN demonstrates strong resilience under both conditions. It consistently models the target distribution more accurately and identifies high-reward modes significantly faster than baseline methods. These results emphasize TD-GFN's ability to maintain performance in the presence of noisy and limited data. Appendix~\ref{apd:few} provide additional evaluations under extreme data scarcity.

\textbf{Robustness to Behavior Policy Quality.}\,
We also examine the sensitivity of TD-GFN to the quality of the behavior policy used during data collection. To this end, we construct two offline datasets. The first, denoted \textit{Median}, is collected using a suboptimal GFlowNet trained for only half the number of steps required for convergence. The second, denoted \textit{Bad}, is generated by a GFlowNet trained with an inverted reward function $IR = R^{-0.1}$, while the dataset still contains the true rewards.

\begin{figure}[htbp]
    \centering
    \includegraphics[width=\linewidth]{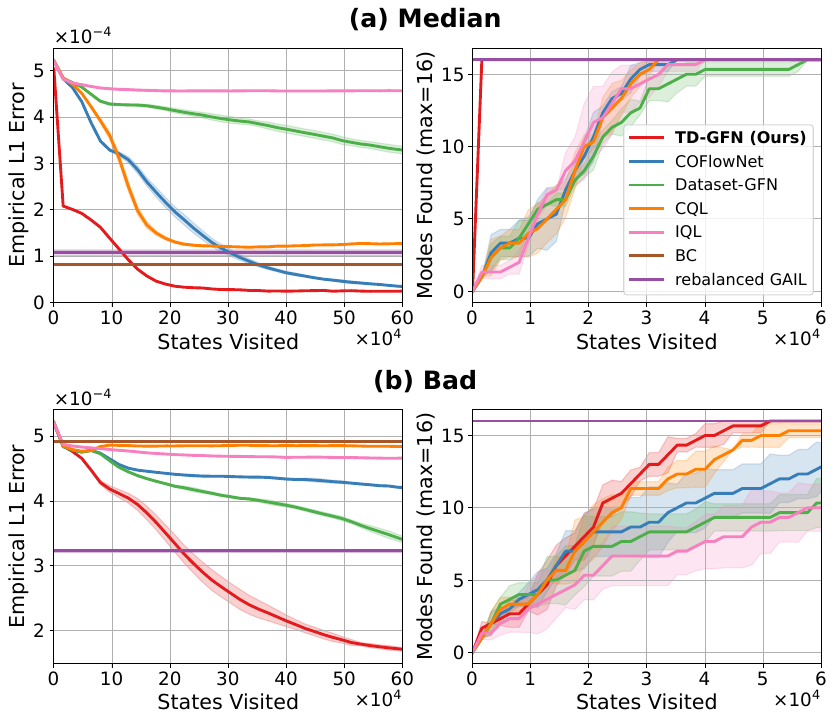}
    \caption{Performance comparison on the $8^4$ Hypergrid task. Policies are trained using the \textit{Median} and \textit{Bad} datasets, each consisting of $1,500$ trajectories.}
    \label{fig:quality}
\end{figure}

As shown in Figure~\ref{fig:quality}, TD-GFN adapts effectively in both scenarios, achieving faster alignment with the target reward distribution and recovering all reward modes earlier than baseline methods. Notably, when the behavior policy diverges substantially from the expert (\textit{Bad}), TD-GFN exhibits significantly more accurate distribution modeling, underscoring its robustness under adverse conditions.

Notably, the $150$ trajectories from the \textit{Mixed} dataset and $1,500$ from the \textit{Bad} dataset cover only $12$ and $6$ out of the $16$ reward modes, respectively. This confirms that our experiments evaluate the model's ability to generate novel states unobserved in the dataset, rather than merely assessing its tendency to overfit the offline data.

\subsection{Biosequence Design}
\label{sec:bioseq}
We next evaluate TD-GFN on a more challenging task involving the design of anti-microbial peptides (AMPs, i.e., short protein sequences) introduced in \citet{jain2022biological}. The experimental setup utilizes two datasets curated from the DBAASP database~\citep{pirtskhalava2021dbaasp}. One dataset is used for training a reward proxy, while the other serves as an oracle to provide ground-truth labels.

Using the proxy reward model provided in \citet{jain2022biological}, we train a Proxy-GFN and compare its performance with TD-GFN, as well as with other baseline methods previously evaluated on the Hypergrid task (Section~\ref{sec:hypergrid}). All methods are trained on the same dataset used for proxy learning, which includes $3,219$ positive AMPs and $4,761$ non-AMPs.

\begin{figure}[htbp]
    \centering
    \includegraphics[width=0.85\linewidth]{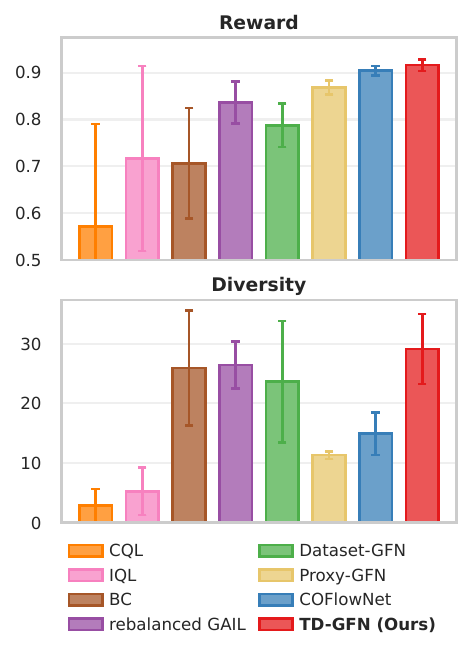}
    \caption{Comparison of reward and diversity among the top $100$ sequences ranked by reward. The \textit{Diversity} metric is defined in Appendix~\ref{apd:bioseq}. All results are reported over three random seeds.
    % All results are averaged over three random seeds, and error bars indicate the standard deviation.
    }
    \label{fig:bioseq}
\end{figure}

We generate $5,000$ sequences using each learned policy and report the reward and diversity metrics for the top $100$ sequences ranked by reward, as shown in Figure~\ref{fig:bioseq}. TD-GFN consistently produces sequences with the highest rewards and greatest diversity. Notably, although rebalanced GAIL merely imitates the policy implied by the rebalanced dataset, it still achieves strong performance—outperforming Dataset-GFN, which is trained directly on the dataset trajectories, in both reward and diversity. These findings highlight the effectiveness of our dataset rebalancing strategy.

\subsection{Molecule Design}
\label{sec:mols}
We further evaluate our method on the real-world molecular design benchmark introduced by \citet{bengio2021flow}. The training dataset simulates historical expert data, comprising $1,500$ trajectories generated by a moderately trained GFlowNet policy (Behavior-GFN). These trajectories are evaluated using the oracle provided in the original study, which was pre-trained on $300,000$ molecules.

\begin{table*}[htbp]
  \caption{Performance comparison on the Molecule Design task. The \textit{Convergence} column denotes the number of trajectories required for performance convergence. Results are reported as mean~$\pm$~standard deviation over three random seeds, with bold values indicating the best performance in each column. The \textit{Dataset} row provides statistics for the $1,500$ samples in the training dataset. Methods positioned above the thin horizontal line are included for reference and are not directly comparable to the main experimental offline baselines.}
  \label{tab:mols}
  \begin{center}
  \begin{tabular}{ccccc}
    \toprule
    Method     & Reward-$10$ (\(\uparrow\))&  Reward-$100$ (\(\uparrow\)) & Reward-$1000$ (\(\uparrow\)) & Convergence (\(\downarrow\))\\
    \midrule
    Dataset & $7.420 \pm 0.088$ & $6.968 \pm 0.226$ & $5.757 \pm 0.635$ & / \\
    Behavior-GFN & $7.534 \pm 0.066$ & $7.220 \pm 0.165$ & $6.504 \pm 0.348$ & / \\
    Oracle-GFN & $7.718 \pm 0.014$ & $7.408 \pm 0.021$ & $6.801 \pm 0.023$ & $44.141 \times 10^4$\\
    \midrule[0.2pt]
    CQL & $7.069$ & $6.643$ & $5.401$ & 0.803 $\times 10^4$ \\
    IQL & $6.902$ & $5.980$ & $4.628$ & $4.104 \times 10^4$ \\
    BraVE & $7.271$ & $6.650$ & $5.590$ & \textbf{0.645} $\bm{\times 10^4}$ \\
    BC & $7.652 \pm 0.053$ & $7.223 \pm 0.035$ & $6.459 \pm 0.016$ & / \\
    GAIL & $7.528 \pm 0.068$ & $7.152 \pm 0.085$ & $6.406 \pm 0.033$ & / \\
    Proxy-GFN& $7.625 \pm 0.063$ & $7.281 \pm 0.067$ & $6.636 \pm 0.097$ &  $43.735 \times 10^4$\\
    Dataset-GFN & $7.550 \pm 0.045$ & $7.198 \pm 0.018$ & $6.474 \pm 0.018$  & $6.030 \times 10^4$\\
    FM-COFlowNet& $7.582 \pm 0.057$ & $7.201 \pm 0.015$ & $6.485 \pm 0.016$ & $5.829 \times 10^4$\\
    QM-COFlowNet& $7.611 \pm 0.020$ & $7.296 \pm 0.022$ & $6.638 \pm 0.010$ & $4.423 \times 10^4$\\
    \textbf{TD-GFN (Ours)} & \textbf{7.733 $\pm$ 0.036} & \textbf{7.450 $\pm$ 0.037} & \textbf{6.810 $\pm$ 0.035} & \textbf{2.749} $\bm{\times 10^4}$\\
    \bottomrule
  \end{tabular}
  \end{center}
\end{table*}

Following the architecture and methodology commonly adopted in prior GFlowNet research~\citep{bengio2021flow,malkin2022trajectory,jang2023learning,tiapkin2024generative}, we train a proxy model on the offline dataset to optimize a GFlowNet baseline termed Proxy-GFN. We also include an Oracle-GFN that utilizes the ground-truth oracle directly as the reward function. This baseline is provided strictly for reference, as it is trained through direct environment interaction rather than the restricted offline dataset. Furthermore, we evaluate both the flow matching (FM) COFlowNet and its quantile-augmented (QM) variant proposed by \citet{zhang2025coflownet}, which employs quantile matching~\citep{zhang2023distributional} to enhance candidate diversity. We also incorporate BraVE~\citep{landers2025brave}, a recent offline RL algorithm designed for discrete action spaces, to extend our comparison to non-GFlowNet methods.

As summarized in Table~\ref{tab:mols}, we report the average top-$k$ rewards from $5,000$ generated samples along with the number of trajectories utilized for loss computation and gradient updates to reach convergence. {This convergence measure reflects both the learning efficiency and the informational utility of the sampled paths for model training. Due to the instability of offline RL under sparse rewards, results for those methods reflect only their best-performing random seed. In this comparison, although BraVE converges fastest in terms of trajectory count, it fails to generate a diverse set of high-reward candidates. This limitation arises from the misalignment between the standard RL paradigm and the diversity objective of the task, which leads to significantly lower rewards than GFlowNet-based or even imitation learning approaches. Excluding offline RL baselines, TD-GFN achieves the best performance in discovering high-reward molecules while requiring the fewest trajectories. Additional comparisons regarding training time and computational resources are provided in Appendix~\ref{apd:time_memo}.

\begin{figure}[htbp]
    \includegraphics[width=\linewidth]{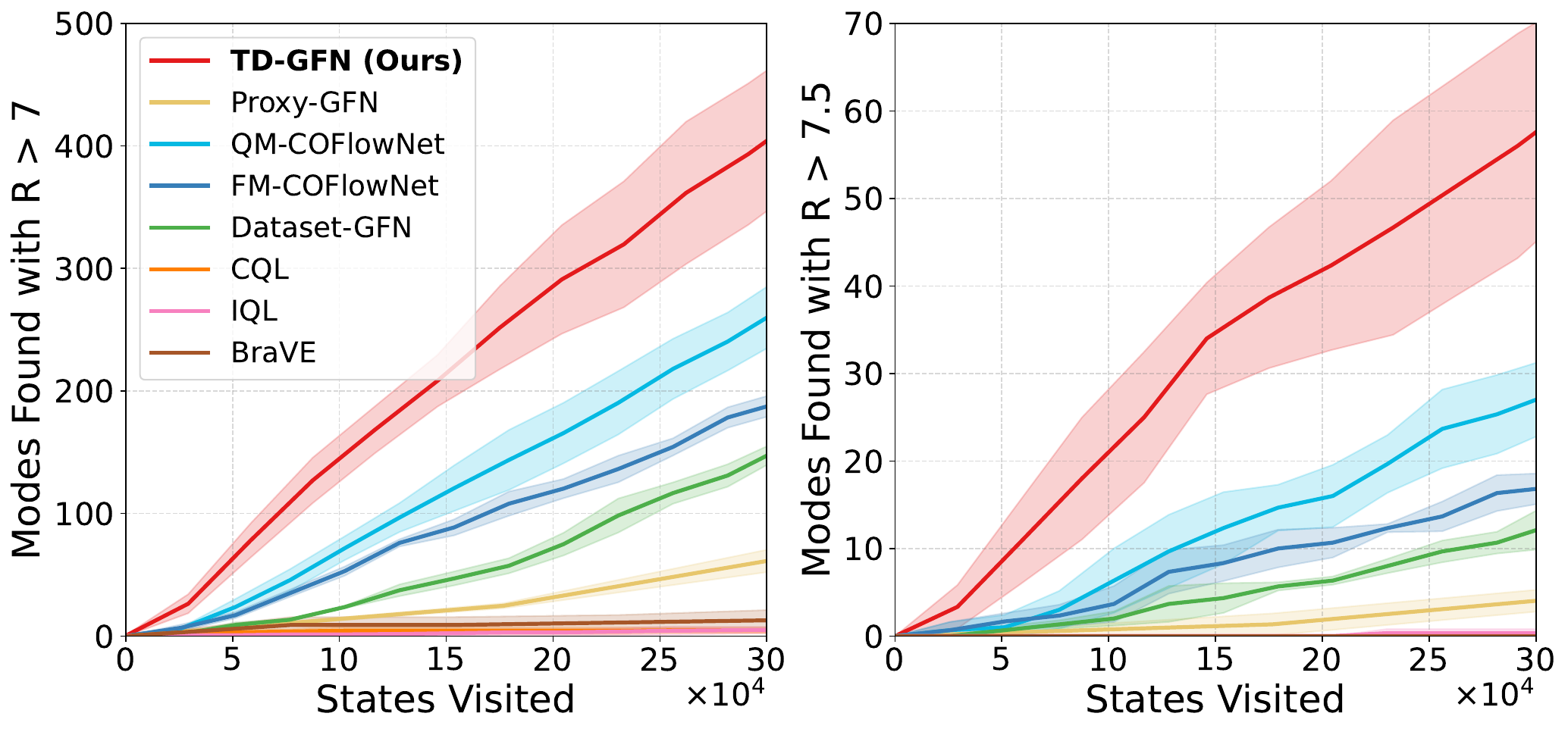}
    \caption{Number of high-reward modes (reward \(>7\) and \(>7.5\)) discovered during training in the Molecule Design task.}
    \label{fig:modes}
\end{figure}

Notably, the powerful intermediate guidance enables TD-GFN to match Oracle-GFN in identifying multiple high-reward molecules while decisively surpassing the behavior policy. This efficiency demonstrates its capacity to transcend training data limitations and suggests strong potential for iterative evolution. Furthermore, TD-GFN converges using $20$ times fewer trajectories. While this results from efficient sampling and the pruned environment, it also highlights a fundamental distinction between training paradigms. Unlike online methods that require extensive trial-and-error because rewards are unknown until a trajectory is complete, offline methods can leverage recorded terminal rewards to prioritize high-utility paths during sampling.

To evaluate molecular diversity, we measure the number of distinct high-reward modes identified by each algorithm. Modes are defined as clusters of high-reward molecules determined by Tanimoto similarity~\citep{bajusz2015tanimoto}. As illustrated in Figure~\ref{fig:modes}, TD-GFN consistently identifies $1.5$ to $2$ times more high-reward modes than the strongest baselines, demonstrating superior exploration capabilities. Notably, despite using DAG pruning, TD-GFN avoids overfitting to narrow regions of the search space. Instead, it guides the policy toward a broader set of high-reward regions by balancing targeted exploitation with diverse exploration. We provide additional details on the Tanimoto similarity among sampled molecules in Appendix~\ref{apd:tanimoto}.

\section{Conclusion}
We presented \textbf{Trajectory-Distilled GFlowNet (TD-GFN)}, a robust proxy-free framework for learning GFlowNets from offline data without requiring out-of-distribution reward queries. By utilizing edge-level guidance learned through inverse reinforcement learning, TD-GFN improves performance and training efficiency via DAG pruning and prioritized backward trajectory sampling.
Extensive experiments across diverse benchmarks demonstrate state-of-the-art results, establishing TD-GFN as a scalable and efficient solution that effectively bridges the gap between limited offline datasets and the complex exploration requirements of generative modeling in real-world applications.

\section*{Acknowledgments}
This work was supported by the National Natural Science Foundation of China Grant 52494974.

\section*{Impact Statement}
TD-GFN targets offline GFlowNet training for compositional generation, with applications in molecule and drug discovery, biological sequence design, and other domains where active reward queries are prohibitive. By improving sample efficiency and diversity from limited offline data, our framework can reduce the number of reward queries needed in costly wet-lab pipelines. The same technical advances, however, can be misused to design harmful molecules or sequences; our work relies on mitigations standard to this line of research, including responsible release of datasets, dual-use review for downstream applications, and domain-specific safety filters at deployment. No other specific societal risks are identified.

\bibliography{refs}

@article{bengio2021flow,
  title={Flow network based generative models for non-iterative diverse candidate generation},
  author={Bengio, Emmanuel and Jain, Moksh and Korablyov, Maksym and Precup, Doina and Bengio, Yoshua},
  journal={Advances in Neural Information Processing Systems},
  volume={34},
  pages={27381--27394},
  year={2021}
}

@article{bengio2023gflownet,
  title={{GFlowNet} foundations},
  author={Bengio, Yoshua and Lahlou, Salem and Deleu, Tristan and Hu, Edward J and Tiwari, Mo and Bengio, Emmanuel},
  journal={Journal of Machine Learning Research},
  volume={24},
  number={210},
  pages={1--55},
  year={2023}
}

@inproceedings{jain2022biological,
  title={Biological sequence design with {GFlowNets}},
  author={Jain, Moksh and Bengio, Emmanuel and Hernandez-Garcia, Alex and Rector-Brooks, Jarrid and Dossou, Bonaventure FP and Ekbote, Chanakya Ajit and Fu, Jie and Zhang, Tianyu and Kilgour, Michael and Zhang, Dinghuai and others},
  booktitle={International Conference on Machine Learning},
  pages={9786--9801},
  year={2022},
  organization={PMLR}
}

@inproceedings{
hu2024amortizing,
title={Amortizing intractable inference in large language models},
author={Edward J Hu and Moksh Jain and Eric Elmoznino and Younesse Kaddar and Guillaume Lajoie and Yoshua Bengio and Nikolay Malkin},
booktitle={The Twelfth International Conference on Learning Representations},
year={2024},
url={https://openreview.net/forum?id=Ouj6p4ca60}
}

@inproceedings{
liu2025efficient,
title={Efficient Diversity-Preserving Diffusion Alignment via Gradient-Informed {GF}lowNets},
author={Zhen Liu and Tim Z. Xiao and Weiyang Liu and Yoshua Bengio and Dinghuai Zhang},
booktitle={The Thirteenth International Conference on Learning Representations},
year={2025},
url={https://openreview.net/forum?id=Aye5wL6TCn}
}

@inproceedings{
zhang2025adversarial,
title={Adversarial Generative Flow Network for Solving Vehicle Routing Problems},
author={Ni Zhang and Jingfeng Yang and Zhiguang Cao and Xu Chi},
booktitle={The Thirteenth International Conference on Learning Representations},
year={2025},
url={https://openreview.net/forum?id=tBom4xOW1H}
}

@article{kumar2020conservative,
  title={Conservative {Q}-learning for offline reinforcement learning},
  author={Kumar, Aviral and Zhou, Aurick and Tucker, George and Levine, Sergey},
  journal={Advances in neural information processing systems},
  volume={33},
  pages={1179--1191},
  year={2020}
}

@inproceedings{kostrikov2021offline,
  title={Offline Reinforcement Learning with Implicit {Q}-Learning},
  author={Kostrikov, Ilya and Nair, Ashvin and Levine, Sergey},
  booktitle={International Conference on Learning Representations ({ICLR})},
  year={2022}
}

@inproceedings{tiapkin2024generative,
  title={Generative flow networks as entropy-regularized rl},
  author={Tiapkin, Daniil and Morozov, Nikita and Naumov, Alexey and Vetrov, Dmitry P},
  booktitle={International Conference on Artificial Intelligence and Statistics},
  pages={4213--4221},
  year={2024},
  organization={PMLR}
}

@article{atanackovic2024investigating,
  title={Investigating generalization behaviours of generative flow networks},
  author={Atanackovic, Lazar and Bengio, Emmanuel},
  journal={arXiv preprint arXiv:2402.05309},
  year={2024}
}

@misc{
wang2023regularized,
title={Regularized Offline {GF}lowNets},
author={Haozhi Wang and Yinchuan Li and yunfeng shao and Jianye HAO},
year={2023},
url={https://openreview.net/forum?id=kbhUUAMZmQT}
}

@inproceedings{
zhang2025coflownet,
title={{COF}lowNet: Conservative Constraints on Flows Enable High-Quality Candidate Generation},
author={Yudong Zhang and Xuan Yu and Xu Wang and Zhaoyang Sun and Chen Zhang and Pengkun Wang and Yang Wang},
booktitle={The Thirteenth International Conference on Learning Representations},
year={2025},
url={https://openreview.net/forum?id=tXUkT709OJ}
}

@article{jang2024pessimistic,
  title={Pessimistic backward policy for {GFlowNets}},
  author={Jang, Hyosoon and Jang, Yunhui and Kim, Minsu and Park, Jinkyoo and Ahn, Sungsoo},
  journal={arXiv preprint arXiv:2405.16012},
  year={2024}
}

@inproceedings{mohammadpour2024maximum,
  title={Maximum entropy {GFlowNets} with soft {Q}-learning},
  author={Mohammadpour, Sobhan and Bengio, Emmanuel and Frejinger, Emma and Bacon, Pierre-Luc},
  booktitle={International Conference on Artificial Intelligence and Statistics},
  pages={2593--2601},
  year={2024},
  organization={PMLR}
}

@article{malkin2210gflownets,
  title={{GFlowNets} and variational inference},
  author={Malkin, Nikolay and Lahlou, Salem and Deleu, Tristan and Ji, Xu and Hu, Edward and Everett, Katie and Zhang, Dinghuai and Bengio, Yoshua},
  journal={URL https://arxiv. org/abs/2210.00580},
  year={2023}
}

@article{
zimmermann2023a,
title={A Variational Perspective on Generative Flow Networks},
author={Heiko Zimmermann and Fredrik Lindsten and Jan-Willem van de Meent and Christian A Naesseth},
journal={Transactions on Machine Learning Research},
issn={2835-8856},
year={2023},
url={https://openreview.net/forum?id=AZ4GobeSLq},
note={}
}

@article{malkin2022trajectory,
  title={Trajectory balance: Improved credit assignment in {GFlowNets}},
  author={Malkin, Nikolay and Jain, Moksh and Bengio, Emmanuel and Sun, Chen and Bengio, Yoshua},
  journal={Advances in Neural Information Processing Systems},
  volume={35},
  pages={5955--5967},
  year={2022}
}

@inproceedings{madan2023learning,
  title={Learning {GFlowNets} from partial episodes for improved convergence and stability},
  author={Madan, Kanika and Rector-Brooks, Jarrid and Korablyov, Maksym and Bengio, Emmanuel and Jain, Moksh and Nica, Andrei Cristian and Bosc, Tom and Bengio, Yoshua and Malkin, Nikolay},
  booktitle={International Conference on Machine Learning},
  pages={23467--23483},
  year={2023},
  organization={PMLR}
}

@inproceedings{
pan2022generative,
title={Generative Augmented Flow Networks},
author={Ling Pan and Dinghuai Zhang and Aaron Courville and Longbo Huang and Yoshua Bengio},
booktitle={The Eleventh International Conference on Learning Representations },
year={2023},
url={https://openreview.net/forum?id=urF_CBK5XC0}
}

@inproceedings{pan2023better,
  title={Better training of {GFlowNets} with local credit and incomplete trajectories},
  author={Pan, Ling and Malkin, Nikolay and Zhang, Dinghuai and Bengio, Yoshua},
  booktitle={International Conference on Machine Learning},
  pages={26878--26890},
  year={2023},
  organization={PMLR}
}

@article{jang2023learning,
  title={Learning energy decompositions for partial inference of {GFlowNets}},
  author={Jang, Hyosoon and Kim, Minsu and Ahn, Sungsoo},
  journal={arXiv preprint arXiv:2310.03301},
  year={2023}
}

@inproceedings{shen2023towards,
  title={Towards understanding and improving {GFlowNet} training},
  author={Shen, Max W and Bengio, Emmanuel and Hajiramezanali, Ehsan and Loukas, Andreas and Cho, Kyunghyun and Biancalani, Tommaso},
  booktitle={International conference on machine learning},
  pages={30956--30975},
  year={2023},
  organization={PMLR}
}

@article{gritsaev2024optimizing,
  title={Optimizing Backward Policies in {GFlowNets} via Trajectory Likelihood Maximization},
  author={Gritsaev, Timofei and Morozov, Nikita and Samsonov, Sergey and Tiapkin, Daniil},
  journal={arXiv preprint arXiv:2410.15474},
  year={2024}
}

@inproceedings{
silva2024on,
title={On Divergence Measures for Training {GF}lowNets},
author={Tiago Silva and Eliezer de Souza da Silva and Diego Mesquita},
booktitle={The Thirty-eighth Annual Conference on Neural Information Processing Systems},
year={2024},
url={https://openreview.net/forum?id=N5H4z0Pzvn}
}

@article{hu2024beyond,
  title={Beyond squared error: Exploring loss design for enhanced training of generative flow networks},
  author={Hu, Rui and Zhang, Yifan and Li, Zhuoran and Huang, Longbo},
  journal={arXiv preprint arXiv:2410.02596},
  year={2024}
}

@article{kim2310local,
  title={Local search {GFlowNets}},
  author={Kim, Minsu and Yun, Taeyoung and Bengio, Emmanuel and Zhang, Dinghuai and Bengio, Yoshua and Ahn, Sungsoo and Park, Jinkyoo},
  journal={URL https://arxiv. org/abs/2310},
  volume={2710},
  year={2024}
}

@article{kim2024adaptive,
  title={Adaptive teachers for amortized samplers},
  author={Kim, Minsu and Choi, Sanghyeok and Yun, Taeyoung and Bengio, Emmanuel and Feng, Leo and Rector-Brooks, Jarrid and Ahn, Sungsoo and Park, Jinkyoo and Malkin, Nikolay and Bengio, Yoshua},
  journal={arXiv preprint arXiv:2410.01432},
  year={2024}
}

@article{lau2024qgfn,
  title={Qgfn: Controllable greediness with action values},
  author={Lau, Elaine and Lu, Stephen and Pan, Ling and Precup, Doina and Bengio, Emmanuel},
  journal={Advances in neural information processing systems},
  volume={37},
  pages={81645--81676},
  year={2024}
}

@inproceedings{madantowards,
  title={Towards Improving Exploration through Sibling Augmented {GFlowNets}},
  author={Madan, Kanika and Lamb, Alex and Bengio, Emmanuel and Berseth, Glen and Bengio, Yoshua},
  booktitle={The Thirteenth International Conference on Learning Representations},
  year={2025}
}

@inproceedings{pan2023stochastic,
  title={Stochastic generative flow networks},
  author={Pan, Ling and Zhang, Dinghuai and Jain, Moksh and Huang, Longbo and Bengio, Yoshua},
  booktitle={Uncertainty in Artificial Intelligence},
  pages={1628--1638},
  year={2023},
  organization={PMLR}
}

@inproceedings{lahlou2023theory,
  title={A theory of continuous generative flow networks},
  author={Lahlou, Salem and Deleu, Tristan and Lemos, Pablo and Zhang, Dinghuai and Volokhova, Alexandra and Hern{\'a}ndez-Garc{\i}a, Alex and Ezzine, L{\'e}na N{\'e}hale and Bengio, Yoshua and Malkin, Nikolay},
  booktitle={International Conference on Machine Learning},
  pages={18269--18300},
  year={2023},
  organization={PMLR}
}

@inproceedings{
chen2024orderpreserving,
title={Order-Preserving {GF}lowNets},
author={Yihang Chen and Lukas Mauch},
booktitle={The Twelfth International Conference on Learning Representations},
year={2024},
url={https://openreview.net/forum?id=VXDPXuq4oG}
}

@inproceedings{brunswic2024theory,
  title={A theory of non-acyclic generative flow networks},
  author={Brunswic, Leo and Li, Yinchuan and Xu, Yushun and Feng, Yijun and Jui, Shangling and Ma, Lizhuang},
  booktitle={Proceedings of the AAAI Conference on Artificial Intelligence},
  volume={38},
  pages={11124--11131},
  year={2024}
}

@article{jain2023gflownets,
  title={{GFlowNets} for ai-driven scientific discovery},
  author={Jain, Moksh and Deleu, Tristan and Hartford, Jason and Liu, Cheng-Hao and Hernandez-Garcia, Alex and Bengio, Yoshua},
  journal={Digital Discovery},
  volume={2},
  number={3},
  pages={557--577},
  year={2023},
  publisher={Royal Society of Chemistry}
}

@inproceedings{jain2023multi,
  title={Multi-objective {GFlowNets}},
  author={Jain, Moksh and Raparthy, Sharath Chandra and Hern{\'a}ndez-Garc{\i}a, Alex and Rector-Brooks, Jarrid and Bengio, Yoshua and Miret, Santiago and Bengio, Emmanuel},
  booktitle={International conference on machine learning},
  pages={14631--14653},
  year={2023},
  organization={PMLR}
}

@article{zhu2023sample,
  title={Sample-efficient multi-objective molecular optimization with {GFlowNets}},
  author={Zhu, Yiheng and Wu, Jialu and Hu, Chaowen and Yan, Jiahuan and Hou, Tingjun and Wu, Jian and others},
  journal={Advances in Neural Information Processing Systems},
  volume={36},
  pages={79667--79684},
  year={2023}
}

@article{roy2023goal,
  title={Goal-conditioned {GFlowNets} for controllable multi-objective molecular design},
  author={Roy, Julien and Bacon, Pierre-Luc and Pal, Christopher and Bengio, Emmanuel},
  journal={arXiv preprint arXiv:2306.04620},
  year={2023}
}

@inproceedings{ghari2023generative,
  title={Generative flow networks assisted biological sequence editing},
  author={Ghari, Pouya M and Tseng, Alex and Eraslan, G{\"o}kcen and Lopez, Romain and Biancalani, Tommaso and Scalia, Gabriele and Hajiramezanali, Ehsan},
  booktitle={NeurIPS 2023 Generative {AI} and Biology (GenBio) Workshop},
  year={2023}
}

@inproceedings{
zhang2023let,
title={Let the Flows Tell:  Solving Graph Combinatorial Problems with {GF}lowNets},
author={Dinghuai Zhang and Hanjun Dai and Nikolay Malkin and Aaron Courville and Yoshua Bengio and Ling Pan},
booktitle={Thirty-seventh Conference on Neural Information Processing Systems},
year={2023},
url={https://openreview.net/forum?id=sTjW3JHs2V}
}

@article{DBLP:journals/corr/abs-2403-07041,
  publtype={informal},
  author={Minsu Kim and Sanghyeok Choi and Jiwoo Son and Hyeonah Kim and Jinkyoo Park and Yoshua Bengio},
  title={Ant Colony Sampling with {GFlowNets} for Combinatorial Optimization},
  year={2024},
  cdate={1704067200000},
  journal={CoRR},
  volume={abs/2403.07041},
  url={https://doi.org/10.48550/arXiv.2403.07041}
}

@inproceedings{zhang2022generative,
  title={Generative flow networks for discrete probabilistic modeling},
  author={Zhang, Dinghuai and Malkin, Nikolay and Liu, Zhen and Volokhova, Alexandra and Courville, Aaron and Bengio, Yoshua},
  booktitle={International Conference on Machine Learning},
  pages={26412--26428},
  year={2022},
  organization={PMLR}
}

@inproceedings{deleu2022bayesian,
  title={{Bayesian} structure learning with generative flow networks},
  author={Deleu, Tristan and G{\'o}is, Ant{\'o}nio and Emezue, Chris and Rankawat, Mansi and Lacoste-Julien, Simon and Bauer, Stefan and Bengio, Yoshua},
  booktitle={Uncertainty in Artificial Intelligence},
  pages={518--528},
  year={2022},
  organization={PMLR}
}

@inproceedings{liu2023generative,
  title={Generative flow network for listwise recommendation},
  author={Liu, Shuchang and Cai, Qingpeng and He, Zhankui and Sun, Bowen and McAuley, Julian and Zheng, Dong and Jiang, Peng and Gai, Kun},
  booktitle={Proceedings of the 29th ACM SIGKDD Conference on Knowledge Discovery and Data Mining},
  pages={1524--1534},
  year={2023}
}

@article{yu2024flow,
  title={Flow of Reasoning: Training {LLMs} for Divergent Problem Solving with Minimal Examples},
  author={Yu, Fangxu and Jiang, Lai and Kang, Haoqiang and Hao, Shibo and Qin, Lianhui},
  journal={arXiv preprint arXiv:2406.05673},
  year={2024}
}

@inproceedings{
venkatraman2024amortizing,
title={Amortizing intractable inference in diffusion models for vision, language, and control},
author={Siddarth Venkatraman and Moksh Jain and Luca Scimeca and Minsu Kim and Marcin Sendera and Mohsin Hasan and Luke Rowe and Sarthak Mittal and Pablo Lemos and Emmanuel Bengio and Alexandre Adam and Jarrid Rector-Brooks and Yoshua Bengio and Glen Berseth and Nikolay Malkin},
booktitle={The Thirty-eighth Annual Conference on Neural Information Processing Systems},
year={2024},
url={https://openreview.net/forum?id=gVTkMsaaGI}
}

@inproceedings{ng2000algorithms,
  title={Algorithms for inverse reinforcement learning.},
  author={Ng, Andrew Y and Russell, Stuart and others},
  booktitle={Icml},
  volume={1},
  pages={2},
  year={2000}
}

@article{ho2016generative,
  title={Generative adversarial imitation learning},
  author={Ho, Jonathan and Ermon, Stefano},
  journal={Advances in neural information processing systems},
  volume={29},
  year={2016}
}

@inproceedings{ziebart2008maximum,
  title={Maximum entropy inverse reinforcement learning.},
  author={Ziebart, Brian D and Maas, Andrew L and Bagnell, J Andrew and Dey, Anind K and others},
  booktitle={Aaai},
  volume={8},
  pages={1433--1438},
  year={2008},
  organization={Chicago, IL, USA}
}

@article{luo2022transferable,
  title={Transferable Reward Learning by Dynamics-Agnostic Discriminator Ensemble},
  author={Luo, Fan-Ming and Cao, Xingchen and Qin, Rong-Jun and Yu, Yang},
  journal={arXiv preprint arXiv:2206.00238},
  year={2022}
}

@inproceedings{
luo2024rewardconsistent,
title={Reward-Consistent Dynamics Models are Strongly Generalizable for Offline Reinforcement Learning},
author={Fan-Ming Luo and Tian Xu and Xingchen Cao and Yang Yu},
booktitle={The Twelfth International Conference on Learning Representations},
year={2024},
url={https://openreview.net/forum?id=GSBHKiw19c}
}

@inproceedings{
kostrikov2018discriminator,
title={Discriminator-Actor-Critic: Addressing Sample Inefficiency and Reward Bias in Adversarial Imitation Learning},
author={Ilya Kostrikov and Kumar Krishna Agrawal and Debidatta Dwibedi and Sergey Levine and Jonathan Tompson},
booktitle={International Conference on Learning Representations},
year={2019},
url={https://openreview.net/forum?id=Hk4fpoA5Km},
}

@article{garg2021iq,
  title={Iq-learn: Inverse soft-q learning for imitation},
  author={Garg, Divyansh and Chakraborty, Shuvam and Cundy, Chris and Song, Jiaming and Ermon, Stefano},
  journal={Advances in Neural Information Processing Systems},
  volume={34},
  pages={4028--4039},
  year={2021}
}

@inproceedings{ziebart2010modeling,
author = {Ziebart, Brian D. and Bagnell, J. Andrew and Dey, Anind K.},
title={Modeling interaction via the principle of maximum causal entropy},
year = {2010},
isbn = {9781605589077},
publisher = {Omnipress},
address = {Madison, WI, USA},
booktitle={Proceedings of the 27th International Conference on International Conference on Machine Learning},
pages = {1255–1262},
numpages = {8},
location = {Haifa, Israel},
series = {ICML'10}
}

@article{pirtskhalava2021dbaasp,
  title={DBAASP v3: database of antimicrobial/cytotoxic activity and structure of peptides as a resource for development of new therapeutics},
  author={Pirtskhalava, Malak and Amstrong, Anthony A and Grigolava, Maia and Chubinidze, Mindia and Alimbarashvili, Evgenia and Vishnepolsky, Boris and Gabrielian, Andrei and Rosenthal, Alex and Hurt, Darrell E and Tartakovsky, Michael},
  journal={Nucleic acids research},
  volume={49},
  number={D1},
  pages={D288--D297},
  year={2021},
  publisher={Oxford University Press}
}

@article{bajusz2015tanimoto,
  title={Why is Tanimoto index an appropriate choice for fingerprint-based similarity calculations?},
  author={Bajusz, D{\'a}vid and R{\'a}cz, Anita and H{\'e}berger, K{\'a}roly},
  journal={Journal of cheminformatics},
  volume={7},
  pages={1--13},
  year={2015},
  publisher={Springer}
}

@inproceedings{jin2018junction,
  title={Junction tree variational autoencoder for molecular graph generation},
  author={Jin, Wengong and Barzilay, Regina and Jaakkola, Tommi},
  booktitle={International conference on machine learning},
  pages={2323--2332},
  year={2018},
  organization={PMLR}
}

@article{zhang2023distributional,
  title={Distributional {GFlowNets} with Quantile Flows},
  author={Zhang, Dinghuai and Pan, Ling and Chen, Ricky T.~Q. and Courville, Aaron and Bengio, Yoshua},
  journal={Transactions on Machine Learning Research ({TMLR})},
  year={2024}
}

@inproceedings{gilmer2017neural,
  title={Neural message passing for quantum chemistry},
  author={Gilmer, Justin and Schoenholz, Samuel S and Riley, Patrick F and Vinyals, Oriol and Dahl, George E},
  booktitle={International conference on machine learning},
  pages={1263--1272},
  year={2017},
  organization={PMLR}
}

@article{ziegler2019fine,
  title={Fine-tuning language models from human preferences},
  author={Ziegler, Daniel M and Stiennon, Nisan and Wu, Jeffrey and Brown, Tom B and Radford, Alec and Amodei, Dario and Christiano, Paul and Irving, Geoffrey},
  journal={arXiv preprint arXiv:1909.08593},
  year={2019}
}

@article{chen2024opportunities,
  title={On the opportunities and challenges of offline reinforcement learning for recommender systems},
  author={Chen, Xiaocong and Wang, Siyu and McAuley, Julian and Jannach, Dietmar and Yao, Lina},
  journal={ACM Transactions on Information Systems},
  volume={42},
  number={6},
  pages={1--26},
  year={2024},
  publisher={ACM New York, NY, USA}
}

@article{dalrymple2024towards,
  title={Towards guaranteed safe ai: A framework for ensuring robust and reliable ai systems},
  author={Dalrymple, David and Skalse, Joar and Bengio, Yoshua and Russell, Stuart and Tegmark, Max and Seshia, Sanjit and Omohundro, Steve and Szegedy, Christian and Goldhaber, Ben and Ammann, Nora and others},
  journal={arXiv preprint arXiv:2405.06624},
  year={2024}
}

@article{ouyang2022training,
  title={Training language models to follow instructions with human feedback},
  author={Ouyang, Long and Wu, Jeffrey and Jiang, Xu and Almeida, Diogo and Wainwright, Carroll and Mishkin, Pamela and Zhang, Chong and Agarwal, Sandhini and Slama, Katarina and Ray, Alex and others},
  journal={Advances in neural information processing systems},
  volume={35},
  pages={27730--27744},
  year={2022}
}

@inproceedings{xie2024advdiffuser,
  title={AdvDiffuser: Generating Adversarial Safety-Critical Driving Scenarios via Guided Diffusion},
  author={Xie, Yuting and Guo, Xianda and Wang, Cong and Liu, Kunhua and Chen, Long},
  booktitle={2024 IEEE/RSJ International Conference on Intelligent Robots and Systems (IROS)},
  pages={9983--9989},
  year={2024},
  organization={IEEE}
}

@article{zahavy2018learn,
  title={Learn what not to learn: Action elimination with deep reinforcement learning},
  author={Zahavy, Tom and Haroush, Matan and Merlis, Nadav and Mankowitz, Daniel J and Mannor, Shie},
  journal={Advances in neural information processing systems},
  volume={31},
  year={2018}
}

@article{zhang2020automatic,
  title={Automatic curriculum learning through value disagreement},
  author={Zhang, Yunzhi and Abbeel, Pieter and Pinto, Lerrel},
  journal={Advances in Neural Information Processing Systems},
  volume={33},
  pages={7648--7659},
  year={2020}
}

@inproceedings{
hong2023harnessing,
title={Harnessing Mixed Offline Reinforcement Learning Datasets via Trajectory Weighting},
author={Zhang-Wei Hong and Pulkit Agrawal and Remi Tachet des Combes and Romain Laroche},
booktitle={The Eleventh International Conference on Learning Representations },
year={2023},
url={https://openreview.net/forum?id=OhUAblg27z}
}

@inproceedings{cao2024offline,
  title={Offline goal-conditioned reinforcement learning for safety-critical tasks with recovery policy},
  author={Cao, Chenyang and Yan, Zichen and Lu, Renhao and Tan, Junbo and Wang, Xueqian},
  booktitle={2024 IEEE International Conference on Robotics and Automation (ICRA)},
  pages={2838--2844},
  year={2024},
  organization={IEEE}
}

@inproceedings{depeweg2018decomposition,
  title={Decomposition of uncertainty in {Bayesian} deep learning for efficient and risk-sensitive learning},
  author={Depeweg, Stefan and Hernandez-Lobato, Jose-Miguel and Doshi-Velez, Finale and Udluft, Steffen},
  booktitle={International conference on machine learning},
  pages={1184--1193},
  year={2018},
  organization={PMLR}
}

@article{morozov2025revisiting,
  title={Revisiting non-acyclic {GFlowNets} in discrete environments},
  author={Morozov, Nikita and Maksimov, Ian and Tiapkin, Daniil and Samsonov, Sergey},
  journal={arXiv preprint arXiv:2502.07735},
  year={2025}
}

@inproceedings{
fu2018learning,
title={Learning Robust Rewards with Adverserial Inverse Reinforcement Learning},
author={Justin Fu and Katie Luo and Sergey Levine},
booktitle={International Conference on Learning Representations},
year={2018},
url={https://openreview.net/forum?id=rkHywl-A-},
}

@inproceedings{
Kostrikov2020Imitation,
title={Imitation Learning via Off-Policy Distribution Matching},
author={Ilya Kostrikov and Ofir Nachum and Jonathan Tompson},
booktitle={International Conference on Learning Representations},
year={2020},
url={https://openreview.net/forum?id=Hyg-JC4FDr}
}

@inproceedings{Chandak2019LearningAR,
  title={Learning Action Representations for Reinforcement Learning},
  author={Yash Chandak and Georgios Theocharous and James E. Kostas and Emma Jordan and Philip S. Thomas},
  booktitle={International Conference on Machine Learning},
  year={2019},
  url={https://api.semanticscholar.org/CorpusID:59553460}
}

@article{Liu2020EnergyBasedIL,
  title={Energy-Based Imitation Learning},
  author={Minghuan Liu and Tairan He and Minkai Xu and Weinan Zhang},
  journal={ArXiv},
  year={2020},
  volume={abs/2004.09395},
  url={https://api.semanticscholar.org/CorpusID:215827862}
}

@inproceedings{
silva2025when,
title={When do {GF}lowNets learn the right distribution?},
author={Tiago Silva and Rodrigo Barreto Alves and Eliezer de Souza da Silva and Amauri H Souza and Vikas Garg and Samuel Kaski and Diego Mesquita},
booktitle={The Thirteenth International Conference on Learning Representations},
year={2025},
url={https://openreview.net/forum?id=9GsgCUJtic}
}

@article{Campbell2024GenerativeFO,
  title={Generative Flows on Discrete State-Spaces: Enabling Multimodal Flows with Applications to Protein Co-Design},
  author={Andrew Campbell and Jason Yim and Regina Barzilay and Tom Rainforth and T. Jaakkola},
  journal={ArXiv},
  year={2024},
  url={https://api.semanticscholar.org/CorpusID:267523194}
}

@article{Brunswic2025ErgodicGF,
  title={Ergodic Generative Flows},
  author={Leo Maxime Brunswic and Mateo Clemente and Rui Heng Yang and Adam Sigal and Amir Rasouli and Yinchuan Li},
  journal={ArXiv},
  year={2025},
  volume={abs/2505.03561},
  url={https://api.semanticscholar.org/CorpusID:278339228}
}

@inproceedings{Sendera2024ImprovedOT,
  title={Improved off-policy training of diffusion samplers},
  author={Marcin Sendera and Minsu Kim and Sarthak Mittal and Pablo Lemos and Luca Scimeca and Jarrid Rector-Brooks and Alexandre Adam and Yoshua Bengio and Nikolay Malkin},
  booktitle={Neural Information Processing Systems},
  year={2024},
  url={https://api.semanticscholar.org/CorpusID:267523463}
}

@article{Xu2020ErrorBO,
  title={Error Bounds of Imitating Policies and Environments for Reinforcement Learning},
  author={Tian Xu and Ziniu Li and Yang Yu},
  journal={IEEE Transactions on Pattern Analysis and Machine Intelligence},
  year={2020},
  volume={44},
  pages={6968-6980},
  url={https://api.semanticscholar.org/CorpusID:225040357}
}

@article{Fu2020D4RLDF,
  title={{D4RL}: Datasets for Deep Data-Driven Reinforcement Learning},
  author={Justin Fu and Aviral Kumar and Ofir Nachum and G. Tucker and Sergey Levine},
  journal={ArXiv},
  year={2020},
  volume={abs/2004.07219},
  url={https://api.semanticscholar.org/CorpusID:215827910}
}

@article{Gulcehre2020RLUB,
  title={RL Unplugged: Benchmarks for Offline Reinforcement Learning},
  author={Caglar Gulcehre and Ziyun Wang and Alexander Novikov and Tom Le Paine and Sergio Gomez Colmenarejo and Konrad Zolna and Rishabh Agarwal and Josh Merel and Daniel Jaymin Mankowitz and Cosmin Paduraru and Gabriel Dulac-Arnold and Jerry Zheng Li and Mohammad Norouzi and Matthew W. Hoffman and Ofir Nachum and G. Tucker and Nicolas Manfred Otto Heess and Nando de Freitas},
  journal={ArXiv},
  year={2020},
  volume={abs/2006.13888},
  url={https://api.semanticscholar.org/CorpusID:220042082}
}

@inproceedings{
landers2025brave,
title={Bra{VE}: Offline Reinforcement Learning for Discrete Combinatorial Action Spaces},
author={Matthew Landers and Taylor W. Killian and Hugo Barnes and Thomas Hartvigsen and Afsaneh Doryab},
booktitle={The Thirty-ninth Annual Conference on Neural Information Processing Systems},
year={2025},
url={https://openreview.net/forum?id=Oj5tVkjbHD}
}

@misc{zhu2025flowrlmatchingrewarddistributions,
      title={FlowRL: Matching Reward Distributions for {LLM} Reasoning}, 
      author={Xuekai Zhu and Daixuan Cheng and Dinghuai Zhang and Hengli Li and Kaiyan Zhang and Che Jiang and Youbang Sun and Ermo Hua and Yuxin Zuo and Xingtai Lv and Qizheng Zhang and Lin Chen and Fanghao Shao and Bo Xue and Yunchong Song and Zhenjie Yang and Ganqu Cui and Ning Ding and Jianfeng Gao and Xiaodong Liu and Bowen Zhou and Hongyuan Mei and Zhouhan Lin},
      year={2025},
      eprint={2509.15207},
      archivePrefix={arXiv},
      primaryClass={cs.LG},
      url={https://arxiv.org/abs/2509.15207}, 
}
\bibliographystyle{icml2026}
%%%%%%%%%%%%%%%%%%%%%%%%%%%%%%%%%%%%%%%%%%%%%%%%%%%%%%%%%%%%

\newpage
\appendix
\onecolumn

\section{Algorithm Pseudocode}
\label{apd:code}

\begin{algorithm}[ht]
\caption{Trajectory-Distilled GFlowNet (TD-GFN)}
\label{alg:full}
\begin{algorithmic}[1]
\STATE {\bfseries Input:} Offline dataset $\mathcal{D} = \{\tau_i = (s_0, \dots, s_{T_i}, R(s_{T_i}))\}_{i=1}^M$; number of training iterations $N$; batch size $B$; GFlowNet learning rate $\eta$.

\STATE \textbf{Phase 1: Edge Reward Extraction}
\STATE Learn edge rewards $R_E(s, s')$ and the imitation policy $\pi_\psi$ via Algorithm~\ref{alg:edge_reward}.

\STATE \textbf{Phase 2: DAG Pruning}
\STATE Collect edge rewards $\mathcal{D}_{R_E}$ using $R_E(s, s')$ and $\pi_\psi$.
\STATE Prune the environment DAG $G = (V, E)$ based on the thresholding rule defined in Equation~\eqref{eq:prune}.
\STATE Remove disconnected edges to obtain the pruned DAG $G' = (V', E')$.

\STATE \textbf{Phase 3: Policy Training}
\STATE Initialize the parameterized GFlowNet policy $\mathcal{P}_F$.
\FOR{$n = 1$ to $N$}
    \STATE Initialize a batch of trajectories $\mathcal{T}=\varnothing$.
    \FOR{$j = 1$ to $B$}
        \STATE Sample a terminal node $x_j \in \mathcal{X}' \cap \{s_{T_i}\}_{i=1}^M$ with probability $P(x) \propto R(x)$.
        \STATE Initialize a trajectory $\tau_j=\varnothing$, $s_{now} = x_j$, $s_{pre} = x_j$.
        \WHILE{$s_{now} \neq s_0$}
            \STATE Sample $s_{pre} \sim \mathcal{P}_B(\cdot|s_{now})$, where $\mathcal{P}_B$ is defined in Equation~\eqref{eq:P_B} over $G'$.
            \STATE $\tau_j = \tau_j \cup \{(s_{pre}\rightarrow s_{now})\}$.
            \STATE $s_{now} = s_{pre}$
        \ENDWHILE
        \STATE $\mathcal{T} = \mathcal{T} \cup \{\tau_{j}\}$.
    \ENDFOR
    \STATE Optimize $\mathcal{P}_F$ over $G'$ with sampled trajectories $\mathcal{T}$.
\ENDFOR

\STATE {\bfseries Output:} Trained GFlowNet policy $\mathcal{P}_F$.
\end{algorithmic}
\end{algorithm}

\begin{algorithm}[H]
\caption{Edge Reward Extraction}
\label{alg:edge_reward}
\begin{algorithmic}[1]
\STATE {\bfseries Input:} Offline dataset $\mathcal{D} = \{\tau_i = (s_0, \dots, s_{T_i}, R(s_{T_i}))\}_{i=1}^M$; number of training iterations $N'$; batch size $B'$; policy learning rate $\alpha$; discriminator learning rate $\beta$.
\STATE Initialize policy $\pi_\psi$ and discriminator $D_\phi: E \rightarrow (0, 1)$ with random parameters $\psi$ and $\phi$.

\FOR{$n = 1$ to $N'$}
    \STATE Initialize a batch of trajectories $\mathcal{T}'=\varnothing$.
    \FOR{$j = 1$ to $B'$}
        \STATE Sample a trajectory $\tau_j \in \mathcal{D}$ with probability $P(\tau_j) \propto R(s_{T_j})$ .
        \STATE $\mathcal{T}'= \mathcal{T}'\cup \{\tau_j\}$.
    \ENDFOR
    \STATE Extract edge transitions $\{(s, s')\}$ from $\mathcal{T}'$ as a minibatch $\overline{E}$.
    \STATE Compute $\nabla_{\phi} L$ on $\overline{E}$; update $\phi \gets \phi - \beta \nabla_{\phi} L$.
    \STATE Compute $\nabla_{\psi} L$ on $\overline{E}$; update $\psi \gets \psi - \alpha \nabla_{\psi} L$ or apply a policy gradient method.
\ENDFOR

\STATE Extract the edge reward function $R_E$ according to Equation~\eqref{eq:R_E}.
\STATE {\bfseries Output:} Edge reward function $R_E$; imitation policy $\pi_\psi$.
\end{algorithmic}
\end{algorithm}

\newpage
% =============================================================
% Camera-ready addition: full theoretical appendix.
% Notation follows the paper: P_F, P_B (calligraphic),
% R(x), R_E, r*, F*, G=(V,E), X, s_0, s_f.
% =============================================================

\section{Theoretical Analysis of IRL Edge Rewards and Pruning}
\label{apd:theory}

This appendix provides a formal analysis of the two core mechanisms of TD-GFN: (i) the edge reward learned by IRL and its connection to the expert backward policy, and (ii) the robustness of the reward-guided pruning rule.

\subsection{Setup and Notation}

We work in the $\gamma = \lambda = 1$ entropy-regularized RL regime of Equations~\eqref{eq:new_rew}--\eqref{eq:equiva}. Let $G = (V, E)$ be a finite DAG with root $s_0$, terminal set $\mathcal{X} \subseteq V$, and an added absorbing state $s_f$. Define the extended edge set
\[
  \bar{E} := E \cup \{(x, s_f) : x \in \mathcal{X}\}.
\]

Assume the expert distribution used in the IRL objective (Equation~\eqref{eq:IRL}) is generated by a fixed GFlowNet with forward policy $\mathcal{P}_F^*$, backward policy $\mathcal{P}_B^*$, and positive terminal reward $R : \mathcal{X} \to \mathbb{R}_{>0}$, so that
\[
  \Pr_{\mathcal{P}_F^*}(x) = \frac{R(x)}{Z}, \qquad Z := \sum_{x \in \mathcal{X}} R(x).
\]

Define the \emph{reach probability}
\[
  \nu^*(s) := \Pr_{\mathcal{P}_F^*}(\text{trajectory visits } s),
\]
the \emph{state flow}
\[
  F^*(s) := Z \cdot \nu^*(s),
\]
and the \emph{edge flow}
\[
  F^*(s, s') := F^*(s) \, \mathcal{P}_F^*(s'|s).
\]
Then $F^*(x) = R(x)$ for terminals, and $\mathcal{P}_B^*(s|s') = F^*(s, s') / F^*(s')$.

For an edge reward $r$ on $\bar{E}$, write
\[
  J_r(\pi) := \mathcal{H}(\pi) + \mathbb{E}_\pi\!\left[\sum_{t=0}^\infty r(s_t, s_{t+1})\right].
\]
The tabular population max-causal-entropy IRL problem corresponding to Equation~\eqref{eq:IRL} is
\[
  \min_r \Big(\max_\pi J_r(\pi) - J_r(\mathcal{P}_F^*)\Big).
\]
Throughout, we adopt the standard finite-horizon GFlowNet conventions: (i) every state $s \in V$ is reachable from $s_0$ under $\mathcal{P}_F^*$ (so that $F^*(s) > 0$); (ii) every nonterminal state has at least one outgoing edge in $E$ leading toward $\mathcal{X}$, so each trajectory terminates in finitely many steps at some $x \in \mathcal{X}$, after which the deterministic transition $x \to s_f$ is appended (carrying the boundary reward $r(x, s_f) = \log R(x)$ from Equation~\eqref{eq:new_rew}) and then $s_f$ absorbs the trajectory with no further contribution to $J_r(\pi)$; and (iii) $\mathcal{P}_F^*(s' \mid s) > 0$ for all $(s, s') \in E$ (otherwise we restrict to the support). Under these conventions, the infinite sum in $J_r(\pi)$ is in fact a finite sum: each trajectory accumulates rewards along its (finite) edges in $E$, plus the single boundary term $\log R(x)$ at the terminal transition.

We write $r^*(s, s') := \log \mathcal{P}_B^*(s|s')$ for the \emph{canonical edge reward} and $R_E(s, s')$ for the edge reward learned by IRL (Equation~\eqref{eq:R_E}).

\subsection{IRL Edge Rewards and the Backward Policy}
\label{apd:irl_theory}

\begin{proposition}[Trajectory Reweighting Preserves the Backward Policy]
\label{prop:reweight}
Suppose the raw offline data are sampled from a GFlowNet $(\mathcal{P}_F, \mathcal{P}_B, R)$. Reweight trajectories by a positive terminal factor $w(x)$:
\[
  q_w(\tau) \propto w(x_\tau) \cdot p(\tau),
\]
where $x_\tau$ is the terminal state of $\tau$. Then $q_w$ is exactly the path law of another GFlowNet with the \textbf{same backward policy} $\mathcal{P}_B$ and terminal weight $\widetilde{R}(x) = w(x) R(x)$.

In particular, if Section~\ref{sec:edge_reward} reweights by $w(x) = R(x)$, then the pseudo-expert has terminal weight $R(x)^2$, but its nonterminal canonical reward is still $\log \mathcal{P}_B$.
\end{proposition}

\begin{proof}
For a trajectory $\tau = (s_0, s_1, \dots, s_T = x)$,
\[
  p(\tau) = \prod_{t=0}^{T-1} \mathcal{P}_F(s_{t+1}|s_t).
\]
Using detailed balance $F(s_t) \mathcal{P}_F(s_{t+1}|s_t) = F(s_{t+1}) \mathcal{P}_B(s_t|s_{t+1})$,
\[
  p(\tau)
  = \prod_{t=0}^{T-1} \frac{F(s_{t+1}) \mathcal{P}_B(s_t|s_{t+1})}{F(s_t)}
  = \frac{F(x)}{F(s_0)} \prod_{t=0}^{T-1} \mathcal{P}_B(s_t|s_{t+1}).
\]
Since $F(x) = R(x)$ and $F(s_0) = Z := \sum_x R(x)$,
\[
  p(\tau) = \frac{R(x)}{Z} \prod_{t=0}^{T-1} \mathcal{P}_B(s_t|s_{t+1}).
\]
Therefore
\[
  q_w(\tau)
  = \frac{w(x) R(x)}{\sum_y w(y) R(y)} \prod_{t=0}^{T-1} \mathcal{P}_B(s_t|s_{t+1}),
\]
which is exactly the trajectory law of a GFlowNet with terminal weight $\widetilde{R}(x) = w(x) R(x)$ and the same backward policy $\mathcal{P}_B$.
\end{proof}

\begin{theorem}[The Canonical Edge Reward]
\label{thm:canonical}
Define the canonical reward
\[
  r^*(s, s') =
  \begin{cases}
    \log \mathcal{P}_B^*(s|s'), & (s, s') \in E, \\[1mm]
    \log R(s),                   & s \in \mathcal{X},\; s' = s_f.
  \end{cases}
\]
Then the entropy-regularized RL problem has soft value
\[
  \Phi^*(s) = \log F^*(s), \qquad \Phi^*(s_f) = 0,
\]
and its soft-optimal policy is exactly $\mathcal{P}_F^*$.
\end{theorem}

\begin{proof}
For a terminal $x$, the only outgoing edge is $x \to s_f$, so
\[
  e^{r^*(x, s_f) + \Phi^*(s_f)} = e^{\log R(x)} = R(x) = F^*(x),
\]
hence $\Phi^*(x) = \log F^*(x)$.

For a nonterminal state $s$,
\[
  \sum_{s'} e^{r^*(s, s') + \Phi^*(s')}
  = \sum_{s'} e^{\log \mathcal{P}_B^*(s|s') + \log F^*(s')}
  = \sum_{s'} \mathcal{P}_B^*(s|s') F^*(s').
\]
By detailed balance, $\mathcal{P}_B^*(s|s') F^*(s') = F^*(s) \mathcal{P}_F^*(s'|s)$, so
\[
  \sum_{s'} e^{r^*(s, s') + \Phi^*(s')}
  = F^*(s) \sum_{s'} \mathcal{P}_F^*(s'|s)
  = F^*(s).
\]
Therefore $\Phi^*(s) = \log F^*(s)$.

The induced soft-optimal policy is
\[
  \pi^*(s'|s)
  = e^{r^*(s, s') + \Phi^*(s') - \Phi^*(s)}
  = e^{\log \mathcal{P}_B^*(s|s') + \log F^*(s') - \log F^*(s)}
  = \mathcal{P}_F^*(s'|s),
\]
again by detailed balance.
\end{proof}

\begin{corollary}[Interpretation of the Edge Reward]
\label{cor:interpret}
For every edge $(s, s') \in E$,
\[
  r^*(s, s') = \log \mathcal{P}_B^*(s|s') = \log \frac{F^*(s, s')}{F^*(s')}.
\]
That is, the canonical edge reward is the log conditional incoming-flow score. For a fixed child $s'$, ranking parents by $r^*(s, s')$ is exactly ranking them by their contribution $F^*(s, s')$ to the expert flow at $s'$. Low $r^*(s, s')$ means ``conditioned on reaching $s'$, the expert almost never came through $s$,'' which provides a principled justification for using $R_E$ as a pruning criterion.
\end{corollary}

\begin{theorem}[Complete Characterization of IRL Solutions]
\label{thm:irl_char}
Restrict the IRL search to rewards on $\bar{E}$ whose terminal boundary is fixed to $r(x, s_f) = \log R(x)$ for all $x \in \mathcal{X}$. Then a reward $r$ on $\bar{E}$ is a global minimizer of the population IRL problem (Equation~\eqref{eq:IRL}) if and only if there exists a function $\Phi : V \cup \{s_f\} \to \mathbb{R}$ with
\[
  \Phi(s_f) = 0, \qquad \Phi(x) = \log R(x) \quad \forall\, x \in \mathcal{X},
\]
such that for every nonterminal edge $(s, s') \in E$,
\[
  r(s, s') = \log \mathcal{P}_F^*(s'|s) + \Phi(s) - \Phi(s').
\]
Equivalently, every such IRL-optimal reward has the form
\[
  r(s, s') = r^*(s, s') + u(s) - u(s')
\]
for some potential $u : V \cup \{s_f\} \to \mathbb{R}$ satisfying $u(x) = u(s_f) = 0$ on terminals and the absorbing state.
\end{theorem}

\begin{proof}
Let $\mathcal{L}(r) := \max_\pi J_r(\pi) - J_r(\mathcal{P}_F^*)$. Since the maximization includes $\pi = \mathcal{P}_F^*$, one always has $\mathcal{L}(r) \ge 0$. Therefore $r$ is a global minimizer if and only if $\mathcal{L}(r) = 0$, i.e., $\mathcal{P}_F^*$ is itself optimal for reward $r$.

By standard soft Bellman optimality, $\mathcal{P}_F^*$ is soft-optimal for $r$ if and only if there exists a soft value $\Phi$ such that
\[
  \mathcal{P}_F^*(s'|s) = e^{r(s, s') + \Phi(s') - \Phi(s)}.
\]
Rearranging gives $r(s, s') = \log \mathcal{P}_F^*(s'|s) + \Phi(s) - \Phi(s')$.

For a terminal $x$, the only outgoing edge is $x \to s_f$, so $r(x, s_f) = \Phi(x) - \Phi(s_f) = \Phi(x)$, and fixing the terminal reward to $\log R(x)$ implies $\Phi(x) = \log R(x)$.

Conversely, given any $\Phi$ with those boundary values, define $r$ by $r(s, s') = \log \mathcal{P}_F^*(s'|s) + \Phi(s) - \Phi(s')$. Then for every nonterminal state $s$,
\[
  \sum_{s'} e^{r(s, s') + \Phi(s')}
  = \sum_{s'} e^{\log \mathcal{P}_F^*(s'|s) + \Phi(s)}
  = e^{\Phi(s)},
\]
so $\Phi$ satisfies the soft Bellman equation, and the induced soft-optimal policy is $\mathcal{P}_F^*(s'|s)$. Hence $\mathcal{L}(r) = 0$.

Finally, Theorem~\ref{thm:canonical} gives $r^*(s, s') = \log \mathcal{P}_F^*(s'|s) + \log F^*(s) - \log F^*(s')$. Subtracting,
\[
  r(s, s') - r^*(s, s') = \bigl(\Phi(s) - \log F^*(s)\bigr) - \bigl(\Phi(s') - \log F^*(s')\bigr).
\]
Setting $u(s) := \Phi(s) - \log F^*(s)$, since $F^*(x) = R(x)$ and $\Phi(x) = \log R(x)$, we have $u(x) = 0$; also $u(s_f) = 0$. Thus $r = r^* + u(s) - u(s')$.
\end{proof}

\begin{theorem}[Unique Recovery under the Canonical Gauge]
\label{thm:unique}
Among all IRL-optimal rewards (in the sense of Theorem~\ref{thm:irl_char}, i.e., with the fixed terminal boundary $r(x, s_f) = \log R(x)$), the unique one additionally satisfying the \textbf{parent-normalization gauge}
\[
  \sum_{p \in \mathrm{Pa}(s')} e^{r(p, s')} = 1 \qquad \forall\, s' \in V \setminus \{s_0\}
\]
is exactly
\[
  r^*(s, s') = \log \mathcal{P}_B^*(s|s').
\]
Note that $r^*$ is already parent-normalized: $\sum_{p \in \mathrm{Pa}(s')} e^{r^*(p, s')} = \sum_p \mathcal{P}_B^*(p \mid s') = 1$. Hence the parent softmax in Equation~\eqref{eq:P_B} is the identity when applied to $r^*$, and $\hat{\mathcal{P}}_B = \mathcal{P}_B^*$ exactly. For an arbitrary IRL-optimal reward $r = r^* + u(s) - u(s')$ with nonzero potential, the parent softmax in Equation~\eqref{eq:P_B} produces $\hat{\mathcal{P}}_B(p \mid s') \propto \mathcal{P}_B^*(p \mid s') \, e^{u(p)}$, which equals $\mathcal{P}_B^*$ only when $u$ is constant over $\mathrm{Pa}(s')$ for every $s'$. Thus, when the IRL estimator $R_E$ approximates $r^*$ in sup-norm, Equation~\eqref{eq:P_B} approximates $\mathcal{P}_B^*$ with the quantitative TV bound given in Theorem~\ref{thm:consistency}.
\end{theorem}

\noindent\textbf{Application to the rebalanced dataset.}\, The IRL formulation in Equation~\eqref{eq:IRL} solved by TD-GFN treats the rebalanced trajectory distribution $\widetilde{\mathcal{D}}$ as the expert. By Proposition~\ref{prop:reweight}, this rebalanced expert is the GFlowNet with the same backward policy $\mathcal{P}_B^*$ but with terminal weight $\widetilde{R}(x) = w(x)\, R(x)$. Theorems~\ref{thm:irl_char}--\ref{thm:unique} therefore apply with $\widetilde{R}$ in place of $R$ in the terminal boundary; the canonical edge reward on $E$ is invariant, since $r^*(s, s') = \log \mathcal{P}_B^*(s \mid s')$ depends only on $\mathcal{P}_B^*$ and not on the terminal weight. Rebalancing therefore does not affect the downstream use of $R_E$ in DAG pruning and prioritized sampling, both of which depend on $r^*$ only on nonterminal edges. The original reward-proportional target $p(x) = R(x)/Z$ used in Theorem~\ref{thm:distortion} remains based on the original $R$ -- the distribution that TD-GFN's policy ultimately aims to match -- and is independent of the rebalanced weight used for IRL training.

\begin{remark}[Gauge of GAIL-derived $R_E$]
\label{rem:gauge}
The GAIL-style estimator in Equation~\eqref{eq:R_E} does not automatically place $R_E$ in the parent-normalization gauge. Theorem~\ref{thm:unique} is therefore an interpretive result identifying the canonical IRL solution $r^*$, not a claim that $R_E$ equals $r^*$ exactly. The downstream stability of TD-GFN does not rely on exact gauge canonicalization: the pruning analysis in Section~\ref{apd:pruning_theory} only assumes a sup-norm bound $\|R_E - r^*\|_\infty \le \varepsilon$, and the perturbation bound in Theorem~\ref{thm:consistency} converts this into a TV bound for the backward policy implied by Equation~\eqref{eq:P_B}.
\end{remark}

\begin{proof}
Let $r$ be IRL-optimal and satisfy the parent-normalization gauge. By Theorem~\ref{thm:irl_char}, there exists $\Phi$ such that
\[
  r(p, s') = \log \mathcal{P}_F^*(s'|p) + \Phi(p) - \Phi(s').
\]
Define $G(s) := e^{\Phi(s)}$. Then for any non-root state $s' \in V \setminus \{s_0\}$,
\[
  1 = \sum_{p \in \mathrm{Pa}(s')} e^{r(p, s')}
  = e^{-\Phi(s')} \sum_{p \in \mathrm{Pa}(s')} e^{\Phi(p)} \mathcal{P}_F^*(s'|p),
\]
so
\begin{equation}
  G(s') = \sum_{p \in \mathrm{Pa}(s')} G(p) \, \mathcal{P}_F^*(s'|p).
  \tag{$*$}
\end{equation}

We claim that every positive solution of ($*$) has the form $G(s) = c \cdot \nu^*(s)$ for some constant $c > 0$. This is proved by induction over any topological order of the DAG. At the root, $G(s_0) = c = c \cdot \nu^*(s_0)$ since $\nu^*(s_0) = 1$. If the claim holds for all parents of $s'$, then
\[
  G(s') = \sum_p c \cdot \nu^*(p) \, \mathcal{P}_F^*(s'|p) = c \cdot \nu^*(s'),
\]
because the reach probability of $s'$ is exactly the sum over its parents of parent-reach probability times transition probability.

Now use the terminal boundary. For every terminal $x$,
\[
  G(x) = e^{\Phi(x)} = R(x).
\]
But also $G(x) = c \cdot \nu^*(x)$, and $\nu^*(x) = R(x)/Z$. Hence $R(x) = c \cdot R(x)/Z$, which forces $c = Z$. Therefore $G(s) = Z \cdot \nu^*(s) = F^*(s)$, so $\Phi(s) = \log F^*(s)$, and substituting back,
\[
  r(s, s')
  = \log \mathcal{P}_F^*(s'|s) + \log F^*(s) - \log F^*(s')
  = \log \mathcal{P}_B^*(s|s')
  = r^*(s, s').
\]
Uniqueness follows.
\end{proof}

\begin{theorem}[Perturbation Bound for the Recovered Backward Policy]
\label{thm:consistency}
Let $R_E$ be a gauge-fixed estimator of the canonical reward $r^*$ such that, with probability at least $1 - \delta$,
\[
  \|R_E - r^*\|_\infty \le \varepsilon_N(\delta).
\]
Define the recovered backward policy via Equation~\eqref{eq:P_B}:
\[
  \hat{\mathcal{P}}_B(s|s') = \frac{e^{R_E(s, s')}}{\sum_{p \in \mathrm{Pa}(s')} e^{R_E(p, s')}}.
\]
Then, with the same probability,
\[
  \sup_{s' \in V \setminus \{s_0\}}
  \mathrm{TV}\!\left(\hat{\mathcal{P}}_B(\cdot|s'),\; \mathcal{P}_B^*(\cdot|s')\right)
  \le \frac{e^{2\varepsilon_N(\delta)} - 1}{2}.
\]
Also, for any threshold $\tau$,
\[
  r^*(e) \ge \tau + \varepsilon_N(\delta) \;\Longrightarrow\; R_E(e) \ge \tau, \qquad
  r^*(e) \le \tau - \varepsilon_N(\delta) \;\Longrightarrow\; R_E(e) \le \tau.
\]
Hence if $\varepsilon_N(\delta) \to 0$ as $N \to \infty$, then $R_E \to r^* = \log \mathcal{P}_B^*$ and $\hat{\mathcal{P}}_B \to \mathcal{P}_B^*$ uniformly.
\end{theorem}

\begin{proof}
Fix a child $s'$, and write $z_p := r^*(p, s')$, $\hat{z}_p := R_E(p, s')$. By assumption, $|\hat{z}_p - z_p| \le \varepsilon_N$ for all parents $p$. Let
\[
  \mathcal{Z} := \sum_p e^{z_p}, \qquad \hat{\mathcal{Z}} := \sum_p e^{\hat{z}_p}.
\]
Then $e^{-\varepsilon_N} \mathcal{Z} \le \hat{\mathcal{Z}} \le e^{\varepsilon_N} \mathcal{Z}$.

Therefore for each parent $p$,
\[
  \frac{\hat{\mathcal{P}}_B(p|s')}{\mathcal{P}_B^*(p|s')}
  = e^{\hat{z}_p - z_p} \cdot \frac{\mathcal{Z}}{\hat{\mathcal{Z}}}
  \in \left[e^{-2\varepsilon_N},\; e^{2\varepsilon_N}\right].
\]
So $|\hat{\mathcal{P}}_B(p|s') - \mathcal{P}_B^*(p|s')| \le (e^{2\varepsilon_N} - 1) \mathcal{P}_B^*(p|s')$. Summing over parents gives
\[
  \|\hat{\mathcal{P}}_B(\cdot|s') - \mathcal{P}_B^*(\cdot|s')\|_1 \le e^{2\varepsilon_N} - 1,
\]
hence $\mathrm{TV}(\hat{\mathcal{P}}_B(\cdot|s'), \mathcal{P}_B^*(\cdot|s')) \le (e^{2\varepsilon_N} - 1)/2$. Taking the supremum over $s'$ proves the first claim.

The threshold statements are immediate from the sup-norm bound: $R_E(e) \ge r^*(e) - \varepsilon_N$ and $R_E(e) \le r^*(e) + \varepsilon_N$. Uniform convergence follows when $\varepsilon_N \to 0$.
\end{proof}

\subsection{Theoretical Motivation for Pruning}
\label{apd:pruning_theory}

For any score function $g : E \to \mathbb{R}$ and threshold $\tau$, define the pruned graph
\[
  G_\tau(g) = (V, E_\tau(g)), \qquad E_\tau(g) := \{e \in E : g(e) \ge \tau\},
\]
and let $\mathcal{X}_\tau(g) \subseteq \mathcal{X}$ be the terminals reachable from $s_0$ in $G_\tau(g)$.

For each terminal $x \in \mathcal{X}$, let $\mathcal{P}(x) \neq \varnothing$ denote the set of all directed paths from $s_0$ to $x$ (by the reachability assumption in Section~\ref{apd:irl_theory} this set is nonempty for every terminal). Define the \emph{bottleneck score}
\[
  \beta_g(x) := \max_{P \in \mathcal{P}(x)} \min_{e \in P} g(e).
\]
We also write $\mathcal{C}(x)$ for the (finite) family of \emph{$s_0$-to-$x$ edge cuts}, where an edge cut $C \subseteq E$ is a finite subset of edges that intersects every path $P \in \mathcal{P}(x)$.

When the score function is the canonical edge reward $r^*$, we write $\beta(x) := \beta_{r^*}(x)$ for brevity.

\begin{theorem}[Exact Characterization of Hard Pruning]
\label{thm:exact_prune}
For any score function $g$ and any threshold $\tau$,
\[
  x \in \mathcal{X}_\tau(g) \iff \beta_g(x) \ge \tau.
\]
Equivalently,
\[
  \beta_g(x) = \min_{C \in \mathcal{C}(x)} \max_{e \in C} g(e).
\]
\end{theorem}

\begin{proof}
The first equivalence is immediate: $x \in \mathcal{X}_\tau(g)$ means that there exists a path $P \in \mathcal{P}(x)$ such that every edge on $P$ satisfies $g(e) \ge \tau$. This is equivalent to $\exists\, P \in \mathcal{P}(x)$ with $\min_{e \in P} g(e) \ge \tau$, which is exactly $\beta_g(x) \ge \tau$.

For the cut form, fix any path $P \in \mathcal{P}(x)$ and any cut $C \in \mathcal{C}(x)$. Since $P$ must cross $C$, there exists $e \in P \cap C$. Hence
\[
  \min_{e' \in P} g(e') \le g(e) \le \max_{e'' \in C} g(e'').
\]
Taking the maximum over all $P$ and then the minimum over all $C$ gives $\beta_g(x) \le \min_{C \in \mathcal{C}(x)} \max_{e \in C} g(e)$.

For the reverse inequality, define $a := \min_{C \in \mathcal{C}(x)} \max_{e \in C} g(e)$. Consider the thresholded graph $G_a(g)$. If $x$ were not reachable from $s_0$ in $G_a(g)$, let $A$ be the set of nodes reachable from $s_0$ in $G_a(g)$. Then the frontier
\[
  C_A := \{(u, v) \in E : u \in A,\; v \notin A\}
\]
is an $s_0$-to-$x$ cut. By construction, every edge in $C_A$ has score $< a$; otherwise its head would also be reachable. Thus $\max_{e \in C_A} g(e) < a$, contradicting the definition of $a$. Therefore $x$ is reachable in $G_a(g)$, and the first part implies $\beta_g(x) \ge a$. Combining both inequalities yields the result.
\end{proof}

\begin{theorem}[Robust Preservation under Estimation Error]
\label{thm:robust}
Let $r^*$ be the oracle edge reward and $R_E$ its learned estimate. Suppose $\|R_E - r^*\|_\infty \le \varepsilon$. Then for any threshold $\tau$,
\[
  \{x \in \mathcal{X} : \beta(x) > \tau + \varepsilon\}
  \;\subseteq\; \mathcal{X}_\tau(R_E)
  \;\subseteq\; \{x \in \mathcal{X} : \beta(x) \ge \tau - \varepsilon\}.
\]
In particular, any terminal $x$ with oracle bottleneck margin $\beta(x) - \tau > \varepsilon$ is guaranteed to survive pruning.
\end{theorem}

\begin{proof}
Take any $x$ with $\beta(x) > \tau + \varepsilon$. By definition, there exists a path $P \in \mathcal{P}(x)$ such that $r^*(e) > \tau + \varepsilon$ for all $e \in P$. Therefore $R_E(e) \ge r^*(e) - \varepsilon > \tau$ for all $e \in P$. Hence all edges on $P$ survive thresholding, so $x \in \mathcal{X}_\tau(R_E)$.

Conversely, if $x \in \mathcal{X}_\tau(R_E)$, then there exists a path $P \in \mathcal{P}(x)$ with $R_E(e) \ge \tau$ for all $e \in P$. Thus $r^*(e) \ge R_E(e) - \varepsilon \ge \tau - \varepsilon$ for all $e \in P$, which implies $\beta(x) \ge \tau - \varepsilon$.
\end{proof}

\begin{corollary}[Mode-Basin Preservation]
\label{cor:mode}
A \emph{mode basin} is any nonempty subset of terminals; let $\mathcal{M}_1, \dots, \mathcal{M}_m \subseteq \mathcal{X}$ be pairwise disjoint mode basins. If for every $i$ there exists some $x_i \in \mathcal{M}_i$ such that $\beta(x_i) > \tau + \varepsilon$, then
\[
  \mathcal{M}_i \cap \mathcal{X}_\tau(R_E) \ne \varnothing \quad \text{for all } i = 1, \dots, m.
\]
So pruning cannot eliminate any of these mode basins.
\end{corollary}

\begin{proof}
By Theorem~\ref{thm:robust}, each witness terminal $x_i$ survives pruning, i.e., $x_i \in \mathcal{X}_\tau(R_E)$. Since $x_i \in \mathcal{M}_i$, we have $\mathcal{M}_i \cap \mathcal{X}_\tau(R_E) \ne \varnothing$.
\end{proof}

\begin{theorem}[Reward-Mass Loss and Target-Distribution Distortion]
\label{thm:distortion}
Let $\mathcal{X}' := \mathcal{X}_\tau(R_E)$ denote the set of surviving terminals, and assume $\mathcal{X}' \neq \varnothing$ (equivalently, $\delta_\tau < 1$). Define the lost reward mass and its normalized version:
\[
  \Delta_\tau := \sum_{x \notin \mathcal{X}'} R(x), \qquad
  \delta_\tau := \frac{\Delta_\tau}{Z}.
\]
Then under the assumption of Theorem~\ref{thm:robust},
\[
  \Delta_\tau \le \sum_{x:\; \beta(x) \le \tau + \varepsilon} R(x).
\]
Define the original reward-proportional target $p(x) := R(x)/Z$, and the target on the pruned graph
\[
  p_\tau(x) := \frac{R(x) \cdot \mathbf{1}\{x \in \mathcal{X}'\}}{Z - \Delta_\tau}.
\]
Then:
\begin{enumerate}[label=(\roman*)]
  \item $\|p - p_\tau\|_{\mathrm{TV}} = \delta_\tau \le \mathbb{P}_{x \sim p}[\beta(x) \le \tau + \varepsilon]$, where the first equality is exact and the inequality comes from the reward-mass bound above.
  \item For any surviving terminals $x, y \in \mathcal{X}'$,
  \[
    \frac{p_\tau(x)}{p_\tau(y)} = \frac{p(x)}{p(y)} = \frac{R(x)}{R(y)}.
  \]
  That is, among surviving terminals, relative probabilities are unchanged; pruning only renormalizes. (When $\mathcal{X}' = \varnothing$ the conditional target $p_\tau$ is undefined; the reward-mass bound $\Delta_\tau \le \sum_{\beta(x) \le \tau + \varepsilon} R(x)$ still holds.)
\end{enumerate}
\end{theorem}

\begin{proof}
By Theorem~\ref{thm:robust}, if $x \notin \mathcal{X}'$, then $\beta(x) \le \tau + \varepsilon$. Therefore
\[
  \Delta_\tau = \sum_{x \notin \mathcal{X}'} R(x) \le \sum_{x:\; \beta(x) \le \tau + \varepsilon} R(x).
\]

Let $Z' := Z - \Delta_\tau$. For $x \in \mathcal{X}'$,
\[
  p_\tau(x) = \frac{R(x)}{Z'} = \frac{Z}{Z'} \cdot p(x) = \frac{1}{1 - \delta_\tau} \cdot p(x),
\]
and for $x \notin \mathcal{X}'$, $p_\tau(x) = 0$. Hence for any $x, y \in \mathcal{X}'$,
\[
  \frac{p_\tau(x)}{p_\tau(y)} = \frac{R(x)/Z'}{R(y)/Z'} = \frac{R(x)}{R(y)}.
\]

For total variation, split the sum over $\mathcal{X}'$ and $\mathcal{X} \setminus \mathcal{X}'$:
\begin{align*}
  \sum_{x \in \mathcal{X}'} |p(x) - p_\tau(x)|
  &= \sum_{x \in \mathcal{X}'} p(x) \left|\,1 - \frac{1}{1 - \delta_\tau}\right|
  = (1 - \delta_\tau) \cdot \frac{\delta_\tau}{1 - \delta_\tau}
  = \delta_\tau, \\
  \sum_{x \notin \mathcal{X}'} |p(x) - p_\tau(x)|
  &= \sum_{x \notin \mathcal{X}'} p(x) = \delta_\tau.
\end{align*}
Therefore $\|p - p_\tau\|_{\mathrm{TV}} = \tfrac{1}{2}(\delta_\tau + \delta_\tau) = \delta_\tau$.

The probability bound is the normalized version of the reward-mass bound:
\[
  \delta_\tau = \frac{\Delta_\tau}{Z}
  \le \frac{1}{Z} \sum_{x:\; \beta(x) \le \tau + \varepsilon} R(x)
  = \mathbb{P}_{x \sim p}[\beta(x) \le \tau + \varepsilon]. \qedhere
\]
\end{proof}

\begin{corollary}[Count Bound on Dropped Modes]
\label{cor:count}
Let $\mathcal{M}_1, \dots, \mathcal{M}_m \subseteq \mathcal{X}$ be pairwise disjoint mode basins (subsets of terminals), and let $W_{\min} := \min_i \sum_{x \in \mathcal{M}_i} R(x) > 0$ be the smallest basin reward mass. Then the number of basins completely removed by pruning is at most
\[
  \frac{\Delta_\tau}{W_{\min}} \le \frac{1}{W_{\min}} \sum_{x:\; \beta(x) \le \tau + \varepsilon} R(x).
\]
In particular, if the right-hand side is $< 1$, no full mode basin can disappear.
\end{corollary}

\begin{proof}
If a basin $\mathcal{M}_i$ is completely removed, all its reward mass contributes to $\Delta_\tau$. Since the basins are disjoint, the total mass of removed basins is at least $W_{\min}$ times the number of removed basins.
\end{proof}

\begin{corollary}[High-Reward Modes Survive under a Monotone Oracle Score]
\label{cor:monotone}
Suppose the oracle score $r^*$ satisfies the downstream monotonicity condition: there exists a nondecreasing function $\psi$ such that for every terminal $x$ and every edge $e$ on any path from $s_0$ to $x$,
\[
  r^*(e) \ge \psi(R(x)).
\]
Then $\beta(x) \ge \psi(R(x))$ for all $x \in \mathcal{X}$. Hence if $R(x) \ge R_{\min}$ and $\psi(R_{\min}) > \tau + \varepsilon$, then $x \in \mathcal{X}_\tau(R_E)$. Therefore all terminals in the high-reward set $\mathcal{X}_{\mathrm{high}} := \{x \in \mathcal{X} : R(x) \ge R_{\min}\}$ survive pruning.
\end{corollary}

\begin{proof}
For any path $P \in \mathcal{P}(x)$, the monotonicity assumption gives $r^*(e) \ge \psi(R(x))$ for every $e \in P$. Therefore $\min_{e \in P} r^*(e) \ge \psi(R(x))$. Taking the maximum over $P \in \mathcal{P}(x)$ yields $\beta(x) \ge \psi(R(x))$. The survival claim then follows from Theorem~\ref{thm:robust}.
\end{proof}

\newpage
\section{Related work}
\label{apd:related}
\paragraph{Generative Flow Networks (GFlowNets).} 
GFlowNets~\citep{bengio2021flow, bengio2023gflownet} are a class of generative models designed to sample compositional objects from an unnormalized reward distribution via sequential action selection. They have been theoretically linked to entropy-regularized reinforcement learning~\citep{tiapkin2024generative, mohammadpour2024maximum} and variational inference~\citep{malkin2210gflownets, zimmermann2023a}, offering a principled framework for distribution matching. 

Recent research has focused on improving GFlowNet training by proposing new balance conditions~\citep{malkin2022trajectory, madan2023learning}, enhancing credit assignment through intermediate rewards~\citep{pan2022generative, pan2023better, jang2023learning}, designing backward policies for improved sampling~\citep{jang2024pessimistic, mohammadpour2024maximum, shen2023towards, gritsaev2024optimizing}, refining sampling and resampling strategies~\citep{atanackovic2024investigating, kim2310local, kim2024adaptive, lau2024qgfn, madantowards}, and exploring alternative training objectives~\citep{silva2024on, hu2024beyond}, including energy-based training approaches for diffusion-structured inference~\citep{Campbell2024GenerativeFO}. Extensions to the GFlowNet framework have expanded its applicability to stochastic dynamics~\citep{pan2023stochastic}, continuous action spaces~\citep{lahlou2023theory}, non-acyclic transitions~\citep{brunswic2024theory,morozov2025revisiting}, and implicit reward feedback~\citep{chen2024orderpreserving}. These advances have broadened the scope of GFlowNets, enabling impactful applications in molecule design~\citep{jain2023gflownets, jain2023multi, zhu2023sample, roy2023goal}, biological sequence generation~\citep{jain2022biological, ghari2023generative}, combinatorial optimization~\citep{zhang2025adversarial, zhang2023let, DBLP:journals/corr/abs-2403-07041}, causal inference~\citep{zhang2022generative, deleu2022bayesian}, recommendation systems~\citep{liu2023generative}, and fine-tuning of large language models~\citep{hu2024amortizing, yu2024flow} and diffusion models~\citep{liu2025efficient, venkatraman2024amortizing}.

Our work is most closely related to recent efforts in offline and proxy-free GFlowNet training. An early approach, RO-GFlowNets~\citep{wang2023regularized}, uses offline action probabilities to constrain the learned policy, while COFlowNet~\citep{zhang2025coflownet} penalizes flow on unseen edges to limit policy coverage to dataset-adjacent regions. Both methods aim to improve policy reliability by imposing coarse constraints that merely align policy behavior or coverage more closely with the dataset. However, such constraints impose blunt limitations that hinder generalization and make performance highly sensitive to data quality, as they fail to fully exploit the trajectory-level information contained in the dataset. Other related works include \citet{atanackovic2024investigating}, which explores an ``offline'' setting but primarily focuses on evaluating sampling strategies, and the KL-weakFM loss~\citep{Brunswic2025ErgodicGF}, which enables proxy-free imitation learning. Concurrently, some research has investigated improving offline training with explicit reward signals and novel exploration strategies~\citep{Sendera2024ImprovedOT}, providing complementary perspectives to our pruning-based framework.

Conceptually, our work shares its core insight with \citet{silva2025when}, which introduces Weighted Detailed Balance (WDB) to emphasize transitions with a larger impact on correctness, particularly in online GFlowNets with GNN-based parameterizations. Their analysis demonstrates that imbalances across edges can systematically affect sampling accuracy. Our work shares this motivation but takes a different route: instead of reweighting balance conditions, we employ inverse reinforcement learning (IRL) to estimate edge rewards. These rewards are then used to guide pruning and prioritized sampling in the offline setting. This distinction in methodology makes the two approaches complementary—WDB could potentially enhance offline training, while our pruning and sampling strategies may in turn benefit online methods.

\paragraph{Inverse Reinforcement Learning (IRL).}
Inverse reinforcement learning (IRL) aims to recover a reward function that explains observed expert behavior in a Markov decision process~\citep{ng2000algorithms}. 
Among its variants, maximum causal entropy IRL is particularly noteworthy, as it addresses reward ambiguity by formulating the objective as finding a policy that maximizes entropy while matching the observed expert behavior~\citep{ziebart2008maximum, ziebart2010modeling}. Building on this framework, Generative Adversarial Imitation Learning (GAIL)~\citep{ho2016generative} reinterprets imitation as a distribution matching problem via adversarial training, where a discriminator distinguishes expert from policy-generated trajectories, and its output serves as a learned reward for policy training. GAIL eliminates the need to explicitly solve the inner IRL loop, significantly improving scalability, and has inspired a broad class of Adversarial Imitation Learning (AIL) methods~\citep{fu2018learning, kostrikov2018discriminator, Kostrikov2020Imitation, garg2021iq}. This line of work has since been extended in various directions, including efforts to improve reward generalization and robustness~\citep{Xu2020ErrorBO,luo2022transferable,luo2024rewardconsistent}.
\newpage

\section{Experimental details}
\label{apd:details}
\subsection{Hypergrid}
\label{apd:hypergrid}
The Hypergrid task is a synthetic benchmark introduced in \citet{bengio2021flow} to evaluate the exploration and generalization capabilities of GFlowNet algorithms. It presents a significant challenge due to its high-dimensional, sparse reward landscape.

The environment consists of a \(D\)-dimensional hypercubic grid with side length \(H\), defining a discrete state space of size \(H^D\). Each state corresponds to a coordinate \( x = (x_1, \ldots, x_D) \in \{0, \ldots, H-1\}^D \). The agent starts at the origin \( x = (0, \ldots, 0) \) and can increment any individual coordinate by one at each step. A trajectory terminates when a special stop action is taken, and the reward is assigned based on the final state.

The reward function is designed to produce multiple sharp reward modes near the corners of the hypercube. Formally, the reward at a terminal state \( x \) is defined as:
\[
R(x) = R_0 + R_1 \prod_{d=1}^D \mathbb{I}\left(\left|\frac{x_d}{H} - 0.5\right| > 0.25\right) + R_2 \prod_{d=1}^D \mathbb{I}\left(0.3 < \left|\frac{x_d}{H} - 0.5\right| < 0.4\right),
\]
where \( R_0 < R_1 \ll R_2 \), and \( \mathbb{I}(\cdot) \) denotes the indicator function. The first term, weighted by \( R_1 \), introduces \( 2^D \) high-reward modes near the grid’s corners, while the second term, weighted by \( R_2 \), creates even sparser regions of extremely high reward, thereby increasing the difficulty of discovering them through exploration alone.

Following the most challenging configuration proposed in \citet{bengio2021flow}, we set the environment parameters to \( R_0 = 10^{-3}, R_1 = 0.5, R_2 = 2 \). This configuration results in a highly multi-modal and sparsely rewarded environment, making it well-suited for evaluating the effectiveness of offline training strategies.

\subsection{Biosequence Design}
\label{apd:bioseq}

The Anti-Microbial Peptide (AMP) Design task, introduced in \citet{jain2022biological}, is a realistic sequence generation benchmark aimed at designing short peptide sequences with strong anti-microbial properties.

The agent constructs sequences by selecting amino acids from a fixed vocabulary of $20$ standard residues, with a maximum sequence length of $50$. At each time step, the agent chooses either an amino acid or a special end-of-sequence token, proceeding in a left-to-right manner. As a result, the transition structure forms a tree, and the trajectories in offline datasets can be directly inferred by the terminal states.

Notably, when the environment DAG degenerates into a tree, the prioritized backward sampling strategy in TD-GFN simplifies to sampling trajectories from the dataset in proportion to their terminal rewards. Despite this simplification, TD-GFN remains effective on this task, which presents substantial challenges due to the large combinatorial space of valid sequences (\( \approx 20^{50} \)) and the extreme sparsity of high-reward regions.

In our experiments, we adopt the \textit{Diversity} metric to evaluate algorithmic performance, following the methodology in \citet{jain2022biological}. For a set of biological sequences \( C \), the metric is defined as:
\[
    \text{Diversity}(C) = \frac{\sum_{x_i, x_j \in C} \text{Lev}(x_i, x_j)}{|C|(|C| - 1)},
\]
where \( \text{Lev}(\cdot, \cdot) \) denotes the Levenshtein distance. All environmental parameters are kept consistent with those used in \citet{jain2022biological}.

\subsection{Molecule Design}
\label{apd:mols}

We evaluate our method on a fragment-based molecular generation task targeting protein binding affinity, specifically for soluble epoxide hydrolase (sEH). The goal is to construct candidate molecules with high predicted affinity for the target protein. Molecules are generated sequentially using the junction tree framework introduced in \citet{jin2018junction}, where each action involves selecting an attachment atom and adding a fragment from a predefined vocabulary of $105$ building blocks.

The reward is defined as the negative binding energy between the generated molecule and the target protein, as predicted by an oracle model pre-trained on $300k$ molecules, following the setup in \citet{bengio2021flow}. To balance reward maximization with output diversity, the final reward is modulated using a reward exponent \(\omega\), such that \( R'(x) = R(x)^\omega \), where \( R(x) \) denotes the raw oracle prediction. All environmental parameters, including \(\omega\), are kept consistent with those used in \citet{bengio2021flow}.

\subsection{Policy Model Architectures and Hyperparameters}
\label{apd:para}

Across the experimental tasks presented in this work, we adopt the model architectures listed in Table~\ref{tab:arch} as the backbone structures for the policy networks.

\begin{table}[ht]
    \caption{Policy network architectures used across different experimental tasks.}
    \label{tab:arch}
    \begin{center}
    \begin{tabular}{cc}
    \toprule
    \textbf{Task} & \textbf{Model architecture}\\
    \midrule
    Hypergrid & MLP([input, 256, 256, output])\\
    Biosequence Design & MLP([input, 128, 128, output])\\
    Molecule Design & MPNN~\citep{gilmer2017neural} v4\\
    \bottomrule
  \end{tabular}
  \end{center}
\end{table}

\begin{table}[ht]
    \caption{Hyperparameters used across different tasks and methods.}
    \label{tab:para}
    \begin{center}
    \begin{tabular}{cccc}
    \toprule
    \textbf{Method} & \textbf{Hyperparameter} & \textbf{Task} &  \textbf{Value} \\
    \midrule
    \multirow{5}{*}{All}
    & \multirow{3}{*}{Policy learning rate \(\eta\)}
        & Hypergrid & \(1\times 10^{-5}\) \\
    &   & Molecule Design & \(1\times 10^{-4}\) \\
    &   & Biosequence Design & \(1\times 10^{-3}\) \\

    & \multirow{1}{*}{Entropy coef. \(\lambda\)}
        & All & 0.01 \\
        
    & \multirow{1}{*}{Seeds}
        & All & 0,1,2 \\

    \midrule
    
    \multirow{9}{*}{TD-GFN}
    & \multirow{3}{*}{Pruning coef. \(K\)}
        & Hypergrid & \(7.0\) \\
    &   & Molecule Design & \(1.0\) \\
    &   & Biosequence Design & \(2.0\) \\

    & \multirow{3}{*}{Actor learning rate \(\alpha\)}
        & Hypergrid & \(1\times 10^{-5}\) \\
    &   & Molecule Design & \(1\times 10^{-4}\) \\
    &   & Biosequence Design & \(1\times 10^{-3}\) \\

    & \multirow{3}{*}{Discriminator learning rate \(\beta\)}
        & Hypergrid & \(3\times 10^{-5}\) \\
    &   & Molecule Design & \(3\times 10^{-4}\) \\
    &   & Biosequence Design & \(3\times 10^{-3}\) \\
    
    \midrule

    \multirow{6}{*}{COFlowNet}
    & \multirow{3}{*}{Regularizing coef. 1}
        & Hypergrid & \(1.0\) \\
    &   & Molecule Design & \(1.0\) \\
    &   & Biosequence Design & / \\

    & \multirow{3}{*}{Regularizing coef. 2}
        & Hypergrid & \(1.0\) \\
    &   & Molecule Design & \(10.0\) \\
    &   & Biosequence Design & \(25.0\) \\

    \midrule
    \multirow{3}{*}{IQL}
    & \multirow{3}{*}{Expectile}
        & Hypergrid & \(0.9\) \\
    &   & Molecule Design & \(0.9\) \\
    &   & Biosequence Design & \(0.8\) \\

    \midrule
    BC & Loss & All & Cross Entropy Loss \\
    
    \bottomrule
    \end{tabular}
    \end{center}
\end{table}

Key algorithmic hyperparameters used across different tasks and methods are summarized in Table~\ref{tab:para}. These values were selected through empirical tuning to ensure stable training and fair comparisons among baselines.\footnote{Our implementation of COFlowNet is based on the official open-source repository available at \url{https://github.com/yuxuan9982/COflownet}.} Specifically, the pruning threshold coefficient \(K\) is tuned by selecting the value that yields the highest ratio of retained recorded transitions to retained randomly selected transitions. For the sensitivity analysis with respect to parameter \(K\), please refer to Appendix~\ref{apd:K-sensitivity}.

\newpage
\section{Additional experimental results}

\subsection{Additional Experiments on Real-world Recommendation Datasets}
\label{apd:rec}
To further evaluate TD-GFN on a \textbf{real-world combinatorial problem modeled as a general DAG}, we introduce an evaluation on the listwise recommendation task following \citet{liu2023generative}.

\paragraph{Environment}
The environment is built upon the MovieLens 1M (ML1M) dataset\footnote{\url{https://grouplens.org/datasets/movielens/1m/}}, a standard and widely-used public benchmark for recommendation systems comprising real user-item interactions. The agent is tasked with autoregressively generating a list (or ``slate'') of $K=6$ unique items for a user. With a total item pool of $3,706$, this presents a large-scale combinatorial generation challenge. Following \citet{liu2023generative}, the reward for a generated list $O$ is defined as the average of its item-wise rewards, $\mathcal{R}(u,O)=\frac{1}{K}\sum_{i\in O}\mathcal{R}(u,i)$, where item-wise rewards are mapped from user ratings (clicks, likes, stars) to a scale of $[0, 3]$. 

Crucially, since the utility of a recommendation slate is invariant to the permutation of its items, any specific set of unique items can be generated via $K! = 720$ distinct sequential paths. This characteristic explicitly structures the environment as a non-degenerate DAG, distinguishing it from tree-structured tasks. For our offline setting, we filtered the public dataset to retain users with at least $25$ interactions. We then constructed the training dataset by directly sampling trajectories of length $K=6$ from these users' real-world interaction logs.

\paragraph{Metrics}
We evaluate all methods using three standard metrics from the \citet{liu2023generative} benchmark:
\begin{itemize}
    \item \textbf{Average Reward}: The mean listwise reward over a large batch of generated samples.
    \item \textbf{Max Reward}: The reward of the single best list generated.
    \item \textbf{Coverage}: The total number of unique items that appear across all generated lists, measuring sample diversity.
\end{itemize}

\paragraph{Results}
The performance of all methods on the offline ML1M dataset is presented in Table~\ref{tab:ml1m-results}. TD-GFN consistently outperforms other offline GFN baselines; notably, it achieves reward performance competitive with the Oracle-GFN while exhibiting significantly higher and more stable diversity.
\begin{table}[h]
\caption{Performance comparison on the ML1M listwise recommendation task. Results are reported as mean $\pm$ standard deviation over three random seeds.}
\label{tab:ml1m-results}
\begin{center}
\begin{tabular}{lccc}
\toprule
\textbf{Method} & \textbf{Avg. Reward ($\uparrow$)} & \textbf{Max Reward ($\uparrow$)} & \textbf{Coverage ($\uparrow$)} \\
\midrule
Oracle-GFN & $\textbf{2.092} \pm \textbf{0.033}$ & $\textbf{2.930} \pm \textbf{0.017}$ & $113.7 \pm 72.0$ \\
\midrule[0.2pt]
Dataset-GFN & $1.681 \pm 0.202$ & $2.674 \pm 0.184$ & $7.9 \pm 0.9$ \\
QM-COFlowNet & $1.787 \pm 0.022$ & $2.843 \pm 0.020$ & $119.2 \pm 6.1$ \\
\textbf{TD-GFN (Ours)} & $\textbf{1.988} \pm \textbf{0.006}$ & $\textbf{2.914} \pm \textbf{0.013}$ & $\textbf{186.8} \pm \textbf{6.3}$ \\
\bottomrule
\end{tabular}
\end{center}
\end{table}

\subsection{Algorithm trained on extremely limited data}
\label{apd:few}
To further assess the robustness of TD-GFN under extreme data scarcity, we conduct experiments on the \(H = 8\), \(D = 4\) Hypergrid environment using only \(\mathbf{30}\) training trajectories from the \textit{Expert}/\textit{Mixed} datasets which cover merely $\mathbf{5}$ and $\mathbf{3}$ modes respectively, as illustrated in Figure~\ref{fig:few}. In this setting, algorithms are prone to overfitting due to the limited data. For example, both COFlowNet and CQL exhibit non-monotonic trends in \textit{Empirical L1 Error}, initially decreasing but later increasing during training, indicating overfitting to the limited known regions.

In contrast, TD-GFN consistently converges toward the target distribution throughout training. Owing to the strong generalization capabilities of the learned edge rewards, it avoids overfitting to the dataset and quickly identifies high-reward modes. These results highlight TD-GFN’s ability to maintain a favorable balance between exploration and exploitation, even under severely constrained data conditions.

\begin{figure}[ht]
    \centering
    \includegraphics[width=0.85\linewidth]{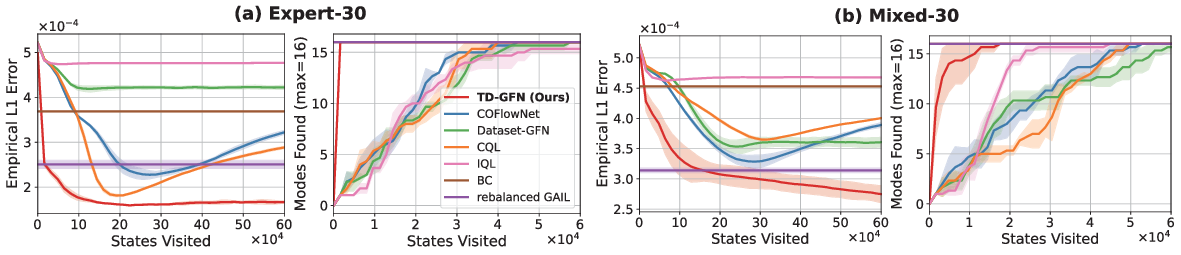}
    \caption{Performance comparison on the \(8^4\) Hypergrid task using $30$ trajectories from the \textit{Expert} and \textit{Mixed} datasets.}
    \label{fig:few}
\end{figure}

\subsection{Algorithm trained on a larger Hypergrid}
\label{apd:larger}
We evaluate the performance of TD-GFN in more challenging settings with sparser rewards, specifically using larger \(20^4\) and \(256^2\) Hypergrid environments. 
For the \(20^4\) task, as shown in Figure~\ref{fig:larger}, TD-GFN achieves strong performance across all four datasets, consistently and significantly outperforming COFlowNet. 
Furthermore, Figure~\ref{fig:256} demonstrates that in the \(256^2\) environment, TD-GFN successfully identifies all $4$ widely separated modes using only $100$ training trajectories. 
Notably, even when trained on the ``Bad'' dataset which contains only $3$ modes, our method effectively generalizes to discover the missing mode.
\begin{figure}[ht]
    \centering
    \includegraphics[width=0.85\linewidth]{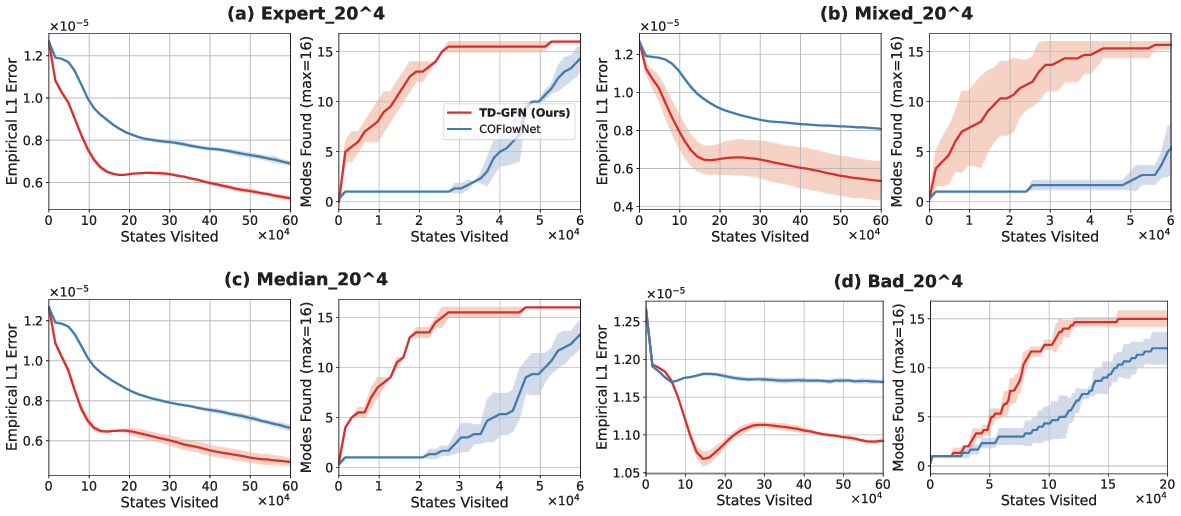}
    \caption{Performance comparison on the \(20^4\) Hypergrid task using $1,500$ trajectories from four types of datasets.}
    \label{fig:larger}
\end{figure}

\begin{figure}[ht]
    \centering
    \includegraphics[width=0.85\linewidth]{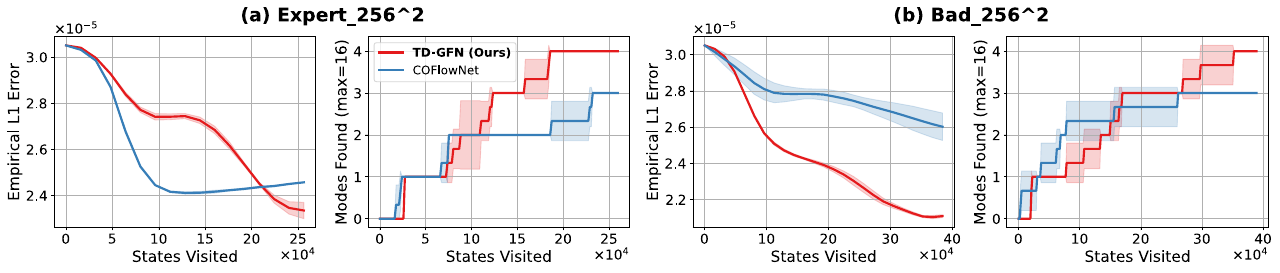}
    \caption{Performance comparison on the \(256^2\) Hypergrid task using $100$ trajectories from four types of datasets.}
    \label{fig:256}
\end{figure}

\subsection{Algorithm Visualization on Hypergrid}
\label{apd:visual}

We visualize the behavior of out proposed TD-GFN in an \(8 \times 8\) Hypergrid environment using a dataset comprising 400 transitions, constructed by combining trajectories collected from an expert policy ($50\%$) and a random policy ($50\%$). The edge rewards learned via IRL using Algorithm~\ref{alg:edge_reward} are shown in Figure~\ref{fig:visual}(b). Compared to the raw edge visitation frequencies from the dataset (Figure~\ref{fig:visual}(a)), the learned edge rewards place greater emphasis on transitions that lead toward high-reward modes.

Notably, the learned edge rewards exhibit strong generalization: they assign meaningful values even to transitions that do not appear in the dataset (e.g., those in the topmost row), thereby guiding the policy toward out-of-distribution high-reward states, such as the modes in the top-right corner.

Figure~\ref{fig:visual}(c) shows the pruned DAG based on the edge rewards in Figure~\ref{fig:visual}(b). As illustrated, the resulting subgraph not only removes many transitions leading to low-reward regions—despite their presence in the dataset—but also constructs clear pathways toward high-reward modes, even when such paths were never encountered during data collection.

\begin{figure}[htbp]
    \centering
    \includegraphics[width=0.8\linewidth]{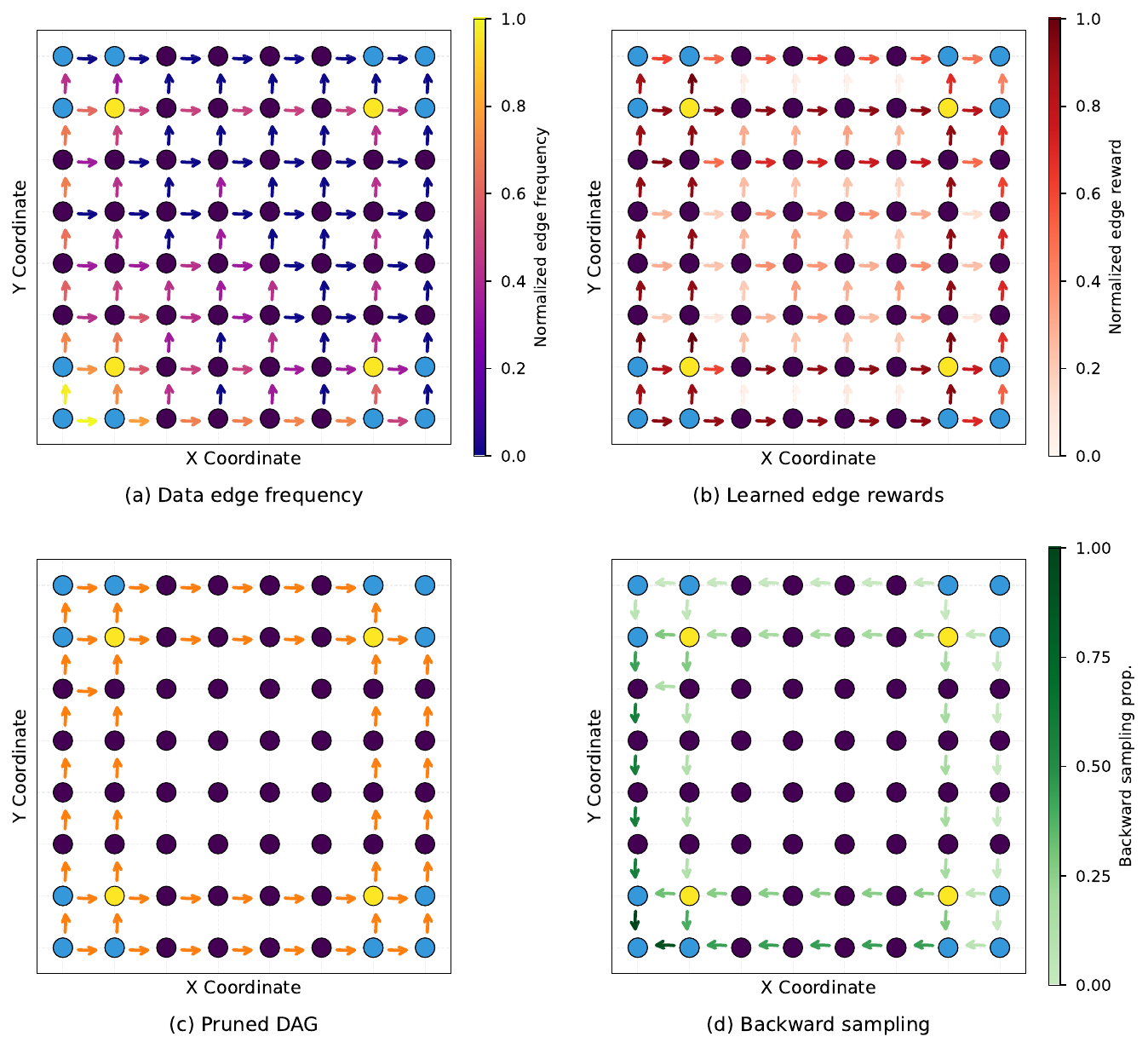}
    \caption{Visualization of TD-GFN in the \(8 \times 8\) Hypergrid environment. (a) Normalized edge visitation frequencies in a dataset composed of $400$ transitions obtained by combining trajectories collected separately from an expert policy ($50\%$) and a random policy ($50\%$); (b) Edge rewards learned via inverse reinforcement learning (IRL); (c) The pruned sub-DAG consisting of high-reward edges (highlighted in yellow) based on edge rewards; (d) The sampling probability of edges under backward policy induced by edge rewards and object rewards.}
    \label{fig:visual}
\end{figure}

\subsection{Sensitivity analysis of the pruning threshold coefficient \(K\)}
\label{apd:K-sensitivity}
In our method, the pruning threshold hyperparameter \(K\) is used to guide edge removal in a statistically principled manner (Equation~\eqref{eq:prune}). It is tuned by selecting the value that yields the highest ratio of retained recorded transitions to retained randomly selected transitions in this work. Our experiments demonstrate that applying a fixed \(K\,(=7)\) across different datasets within the same environment yields strong performance without the need for dataset-specific tuning. While future work may explore more theoretically grounded strategies for DAG pruning, here we provide a sensitivity analysis of this parameter. As shown in Figure~\ref{fig:K-sensitivity}, our method remains robust across different values of \(K\) without catastrophic degradation.

\begin{figure}[htbp]
    \centering
    \includegraphics[width=0.85\linewidth]{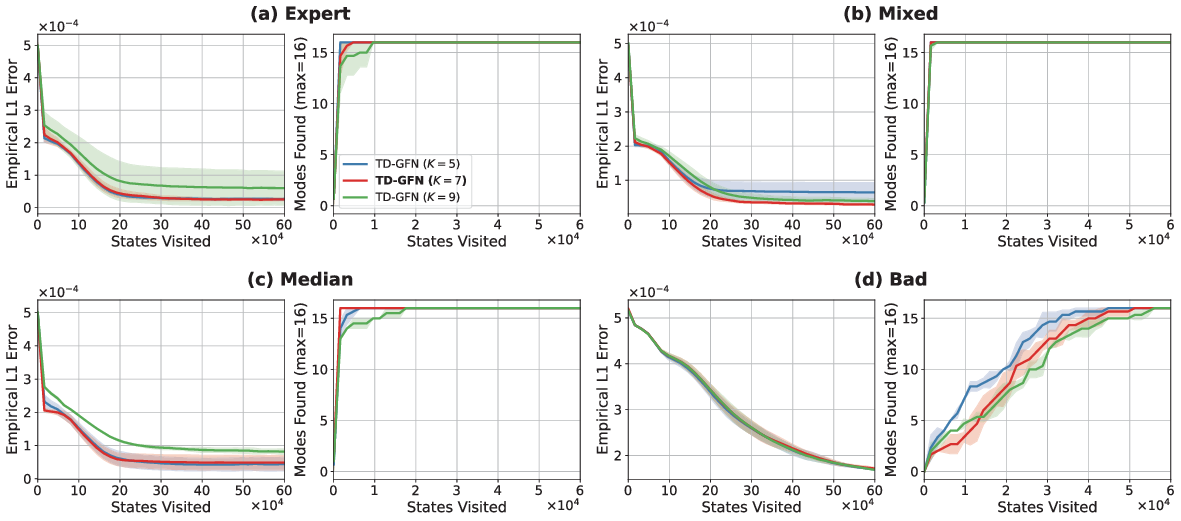}
    \caption{Performance comparison between different values of the pruning coefficient $K$ on the $8^4$ Hypergrid task. Policies are trained using four types of datasets, each consisting of $1,500$ trajectories.}
    \label{fig:K-sensitivity}
\end{figure}

\subsection{Time consumption and resource usage}
\label{apd:time_memo}

For the Molecule Design task (Section~\ref{sec:mols}), all algorithms are trained on an NVIDIA Tesla A100 80GB GPU. We report preprocessing time, per-epoch training time, total convergence time, and peak GPU memory usage for various GFlowNet training methods in Table~\ref{tab:time_memo}.

For Proxy-GFN, preprocessing refers to training the proxy model. For COFlowNet, it involves constructing a dictionary that indexes all child nodes from the dataset for each state. For TD-GFN, preprocessing includes both learning the edge rewards and pruning the environment DAG, corresponding to Phase 1 and Phase 2 in Algorithm~\ref{alg:full}.

\begin{table}[htbp]
  \caption{Training time and GPU memory usage comparison across GFlowNet variants on the Molecule Design task.}
  \label{tab:time_memo}
  \begin{center}
  \begin{tabular}{ccccc}
    \toprule
    \textbf{Method} & \textbf{Preprocessing Time} & \textbf{Epoch Time} & \textbf{Convergence Time} & \textbf{GPU Memory} \\
    \midrule
    Proxy-GFN & 10h & 4.82s & 4h & 22.88G \\
    Dataset-GFN & / & \textbf{0.56s} & 0.47h & 10.08G \\
    FM-COFlowNet   & 2h & 0.76s & 0.63h & 12.81G \\
    QM-COFlowNet   & 2h & 1.16s & 0.71h & 63.80G \\
    TD-GFN (Ours)      & \textbf{0.1h} & \textbf{0.59s} & \textbf{0.20h} & 12.01G \\
    \bottomrule
  \end{tabular}
  \end{center}
\end{table}

As shown, by predicting pruned edges through a single forward pass of a neural network—rather than querying from a pre-constructed dictionary—TD-GFN significantly reduces preprocessing time and incurs minimal computational overhead during training. Furthermore, owing to its rapid convergence (as demonstrated in Table~\ref{tab:mols}), TD-GFN can learn a high-performing GFlowNet policy in under 30 minutes, highlighting its strong time efficiency.

Although the edge reward learning component introduces a modest computational cost, it remains negligible compared to the overhead of more complex mechanisms such as quantile matching. Despite this, TD-GFN outperforms methods enhanced with quantile-based techniques, underscoring its effectiveness and scalability.

\subsection{Ablation Study}
\label{apd:ablation}
In this section, we conduct ablation experiments on the \(8^4\) Hypergrid task to evaluate the contribution of individual components within the proposed TD-GFN framework.

\paragraph{Dataset Rebalancing.} As described in Section~\ref{sec:edge_reward}, we perform dataset rebalancing to adjust the offline data distribution to better approximate that of an expert policy. In Figure~\ref{fig:rebalancing}, we evaluate TD-GFN with and without rebalancing on both the \textit{Expert} and \textit{Bad} datasets. For additional comparison, we also train the offline GFlowNet method COFlowNet  on the rebalanced datasets.

The results show that rebalancing the \textit{Expert} dataset does not degrade policy learning, despite the mild distortion introduced to the original data distribution. More importantly, rebalancing significantly improves policy performance when the dataset is collected under a behavior policy that diverges substantially from the expert.

\begin{figure}[htbp]
    \centering
    \includegraphics[width=0.85\linewidth]{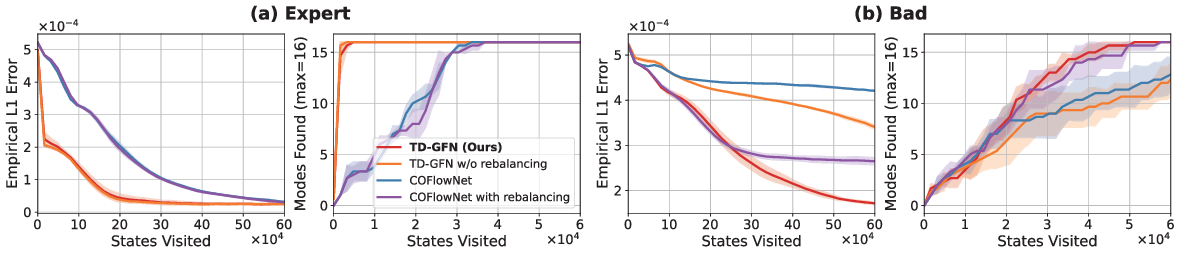}
    \caption{Performance comparison with and without dataset rebalancing on \textit{Expert} and \textit{Bad} datasets, each consisting of \(1,500\) trajectories.}
    \label{fig:rebalancing}
\end{figure}

\paragraph{Pruning and Backward Sampling.} 
To assess the individual impact of TD-GFN’s components, we conduct ablations on the \textit{Mixed} dataset. We compare the full TD-GFN with three variants: (i) a baseline without DAG pruning; (ii) a dataset-pruned variant that removes all edges not observed in the dataset; and (iii) a backward sampling baseline in which trajectories are sampled uniformly in reverse from terminal nodes, without using learned edge rewards. As shown in Figure~\ref{fig:pruning}, both the reward-guided pruning strategy and the prioritized backward sampling procedure are critical to the overall performance of TD-GFN.

\begin{figure}[htbp]
    \centering
    \includegraphics[width=0.5\linewidth]{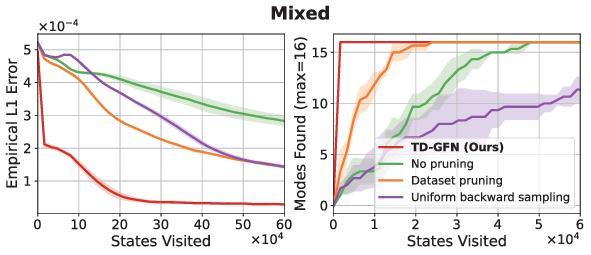}
    \caption{Ablation study on the \textit{Mixed} dataset. We evaluate three TD-GFN variants: (i) no pruning; (ii) dataset-based pruning that removes unseen edges; and (iii) uniform backward sampling from terminal nodes. Each TD-GFN component proves essential for achieving optimal performance.}
    \label{fig:pruning}
\end{figure}

\subsection{Orthogonality to Training Paradigms}
\label{apd:ortho}

As shown in Algorithm~\ref{alg:full}, the improvements introduced by TD-GFN are orthogonal to the specific training paradigm employed after trajectory collection. Due to time and space constraints, our empirical analysis primarily focuses on the compatibility of TD-GFN with existing GFlowNet training objectives, as well as its integration with offline RL methods and the quantile matching technique proposed by \citet{zhang2023distributional}.

\paragraph{Compatibility with GFlowNet training objectives.}
\label{apd:objectives} 
With trajectories sampled on the pruned DAG \(G' = (V', E')\), we can readily apply various GFlowNet training objectives to optimize our policy on this subgraph without requiring substantial modifications, such as the flow matching (FM) objective~\citep{bengio2021flow} and the trajectory balance (TB) objective~\citep{malkin2022trajectory}, as illustrated below.

By setting \(R(s) = 0\) for all non-terminal states \(s\), the FM objective for each sampled trajectory \(\tau\) is given by:
\begin{equation}
\mathcal{L}_{FM}(\tau; \theta) = \sum_{s' \in \tau} \left( \sum_{s:(s \rightarrow s') \in E'} F_\theta(s, s') - R(s') - \sum_{s'':(s' \rightarrow s'') \in E'} F_\theta(s', s'') \right)^2,
\label{eq:fm}
\end{equation}
where \(F_\theta(s, s')\) denotes the flow along edge \((s \rightarrow s')\). The resulting flow-based policy is then defined by sampling child states in proportion to the learned flows at each parent state.

The TB objective for each sampled trajectory $\tau$ is given by:
\begin{equation}
\mathcal{L}_{TB}(\tau; \theta) = \left(Z_\theta + \sum_{(s \rightarrow s') \in E'} \big(\mathcal{P}_F^{G'}(s' \mid s;\theta) - \mathcal{P}_B^{G'}(s \mid s';\theta)\big) - R(x) \right)^2,
\label{eq:tb}
\end{equation}
where $\mathcal{P}_F^{G'}(s' \mid s;\theta)$ and $\mathcal{P}_B^{G'}(s \mid s';\theta)$ denote the forward and backward transition probabilities respectively, both normalized over the edges in the pruned subgraph $G'$ is the reward of the terminal state $x$, and $Z_\theta$ is the learned normalizing constant. Moreover, we find it beneficial to fix $\mathcal{P}_B^{G'}$ to the form given in Equation~\eqref{eq:P_B}, a practice also noted in \citet{malkin2022trajectory}. The results of applying the TB objective on the four datasets used in our main experiments are presented in Figure~\ref{fig:tb}.

\begin{figure}[htbp]
    \centering
    \includegraphics[width=0.85\linewidth]{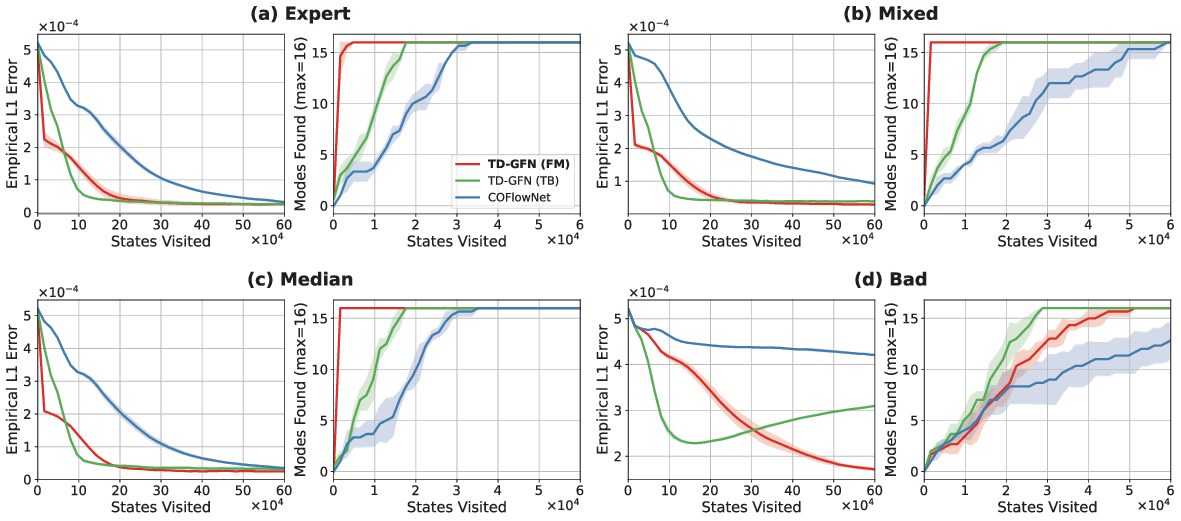}
    \caption{Performance comparison of TD-GFN with different training objectives as well as COFlowNet on the $8^4$ Hypergrid task. Policies are trained using four types of datasets, each consisting of $1,500$ trajectories.}
    \label{fig:tb}
\end{figure}

We observe that the TB objective converges faster than the FM objective, but is less effective in rapidly exploring modes and exhibits weaker stability against overfitting on extreme datasets (e.g., \textit{Bad}). We hypothesize this is because the flow function $F_\theta(s, s')$ in the FM objective can itself be interpreted as an implicit edge-level reward model. From this perspective, our pruning method effectively reshapes the learning landscape for this \textit{flow-reward} model, simplifying its learning task, while our prioritized backward sampling guides it to strategically generalize to high-utility regions. Analogous to how a proxy reward model generalizes across terminal states, our approach enables this flow-reward model to generalize to unseen edges, thereby actively guiding the policy to explore novel states. This synergy might explain the FM objective's enhanced robustness and exploratory capability within our framework. Nevertheless, TD-GFN with the TB objective still substantially outperforms COFlowNet.

\paragraph{Replacing GFlowNet objectives with Offline RL Training.}

Instead of using the standard objectives in GFlowNet training, we examine the effect of replacing it with the Conservative Q-Learning (CQL) objective from offline reinforcement learning~\citep{kumar2020conservative}, resulting in a variant we refer to as TD-CQL. We evaluate TD-CQL on both the \textit{Expert} and \textit{Mixed} datasets, each consisting of \(1,500\) trajectories. Results are presented in Figure~\ref{fig:rl}.

With training components inherited from TD-GFN, TD-CQL successfully discovers all modes using a considerably smaller number of state visits compared to CQL and achieves an \textit{Empirical L1 Error} below \(10^{-4}\) with fewer visits than TD-GFN, demonstrating strong sample efficiency and early-stage accuracy in modeling the target distribution. However, beyond this point, the \textit{Empirical L1 Error} increases, indicating a decline in distributional approximation quality. This degradation arises from the nature of the offline RL objective, which tends to overfit the policy toward only the highest-reward objects, thereby limiting its ability to accurately model the full reward distribution.

\begin{figure}[htbp]
    \centering
    \includegraphics[width=0.85\linewidth]{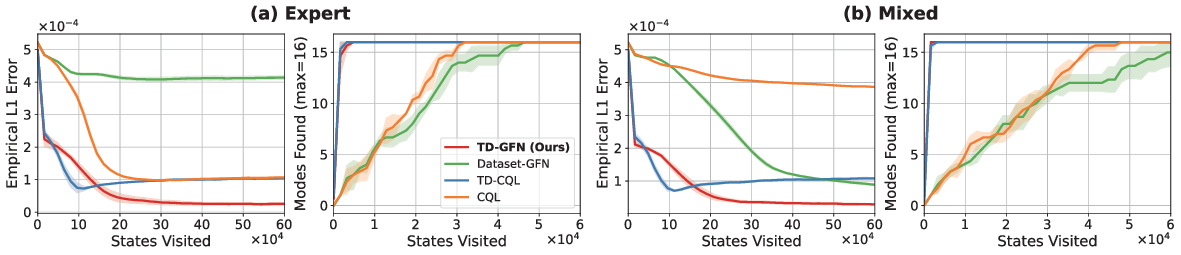}
    \caption{Performance of GFlowNet and CQL with and without our Trajectory-Distilled (TD) framework on the \textit{Expert} and \textit{Mixed} datasets, each consisting of \(1,500\) trajectories.}
    \label{fig:rl}
\end{figure}

\paragraph{Incorporating Quantile Matching.}

Following the approach of \citet{zhang2025coflownet}, we augment TD-GFN with the quantile matching technique proposed in \citet{zhang2023distributional}, resulting in a variant referred to as QM-TD-GFN. We evaluate QM-TD-GFN on the Molecule Design task (Section~\ref{sec:mols}) to assess its impact on the diversity of generated samples.

We observe that incorporating quantile matching into TD-GFN introduces significant training instability. This may be due to the pruned structure of the environment DAG, which reduces the need for a highly uncertain policy. Nevertheless, in cases where training remains stable, QM-TD-GFN is able to discover more high-reward modes than the original TD-GFN, as shown in Figure~\ref{fig:qm}. These findings highlight the potential of quantile matching to enhance the diversity of generated molecules.

\begin{figure}[htbp]
    \centering
    \includegraphics[width=0.5\linewidth]{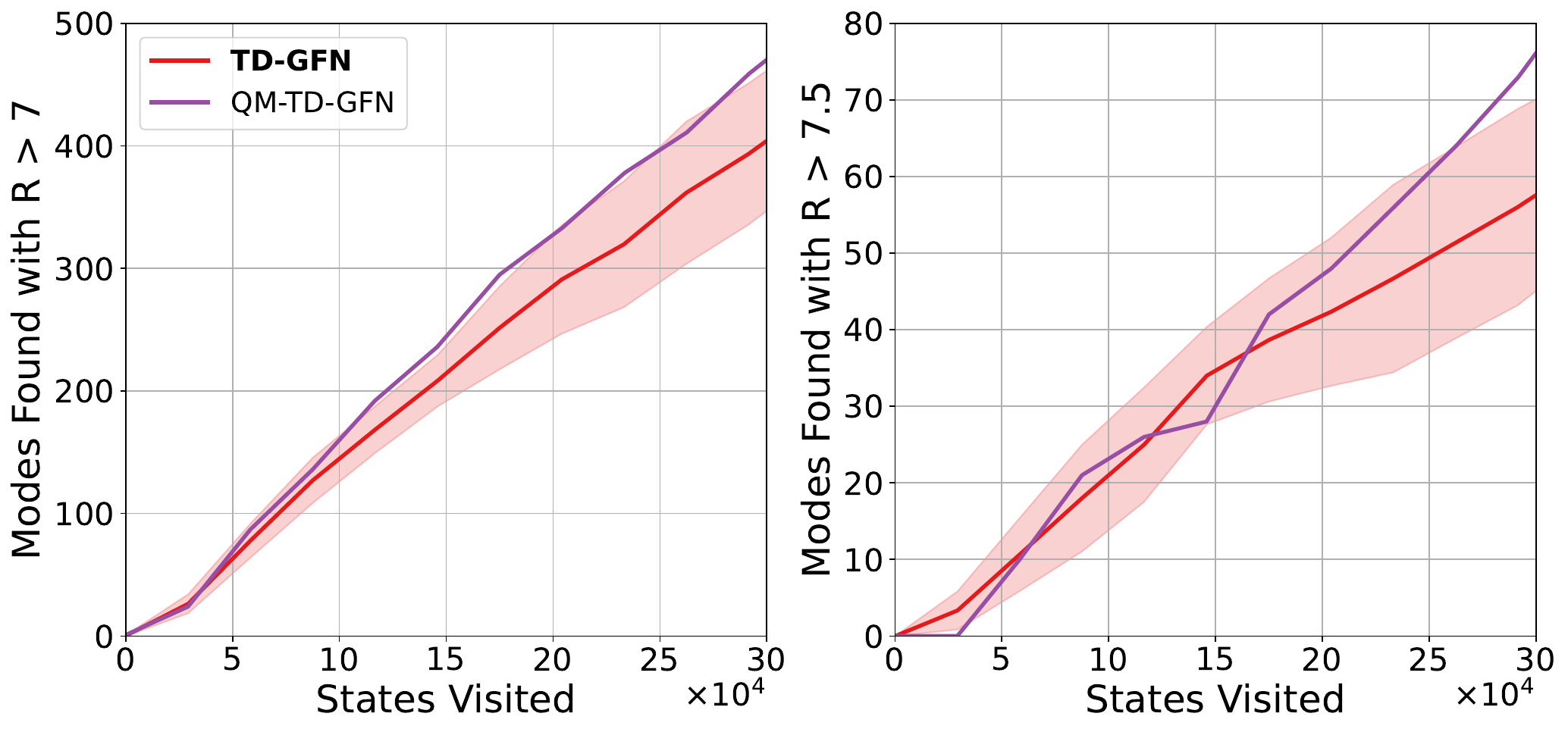}
    \caption{Number of high-reward modes discovered by QM-TD-GFN compared to TD-GFN.}
    \label{fig:qm}
\end{figure}

\subsection{Discussion on Baseline Performance}
\label{apd:baselines}

To gain deeper insight into the behavior of baseline methods, we analyze results on the Hypergrid task, as shown in Figure~\ref{fig:grid_apd}.

\begin{figure}[htbp]
    \centering
    \includegraphics[width=0.85\linewidth]{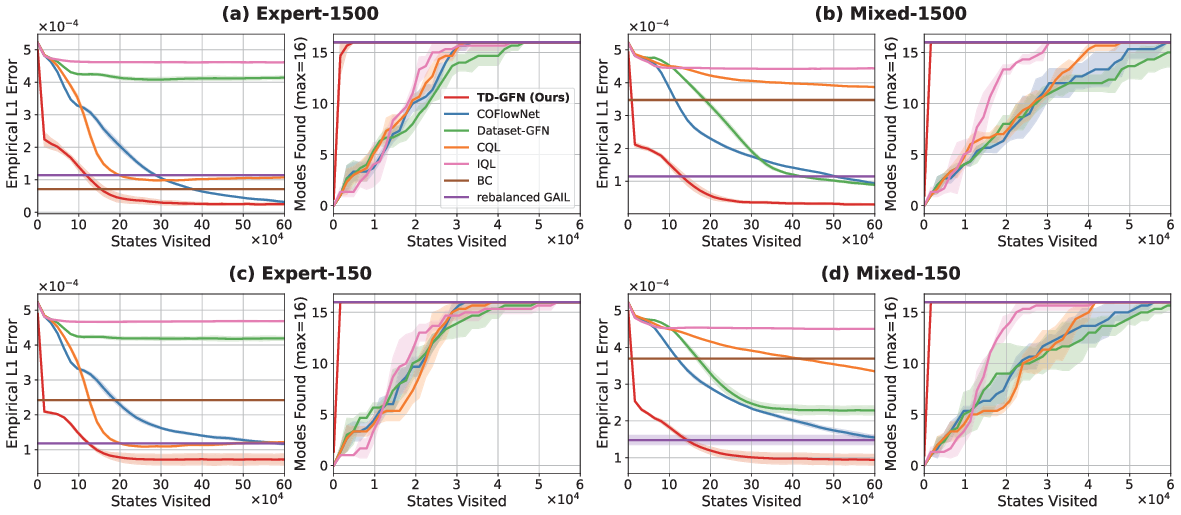}
    \caption{Performance comparison on the \(8^4\) Hypergrid task. The top row uses \(1,500\) trajectories from the \textit{Expert} and \textit{Mixed} datasets for policy training, while the bottom row uses only \(150\) trajectories.}
    \label{fig:grid_apd}
\end{figure}

Across datasets with varying levels of uncertainty, IQL excels at identifying high-reward modes. However, it also exhibits the highest \textit{Empirical L1 Error}, underscoring a core characteristic of offline reinforcement learning algorithms: their objective often prioritizes recovering the most rewarding trajectories over accurately modeling the full reward distribution.

Although CQL is also an offline RL algorithm, its performance differs from IQL due to a distinct training paradigm. Unlike IQL, which learns exclusively from transitions in the dataset, CQL explicitly constrains policy actions to remain close to the data by penalizing the overestimation of Q-values. As a result, its performance is highly sensitive to dataset quality. A similar sensitivity is observed in COFlowNet, which applies conservative regularization to limit its policy coverage to dataset-adjacent regions.

In contrast, GAIL—when trained on a rebalanced dataset—demonstrates stable performance across datasets. It consistently maintains coverage over all modes and achieves an \textit{Empirical L1 Error} slightly above \(1 \times 10^{-4}\). Although GAIL performs slightly worse than Behavior Cloning (BC) in scenarios with abundant expert data, it significantly outperforms BC under noisy or sparse data conditions and achieves performance comparable to COFlowNet. These results highlight the benefit of incorporating reward signals from terminal states during training by rebalancing the dataset.

\subsection{Analysis of Sample Diversity in Molecule Design}
\label{apd:tanimoto}
In the main paper (Figure~\ref{fig:modes}), we report sample diversity using the number of high-reward modes discovered, which is a task-oriented metric evaluating the intersection of exploration and exploitation. To provide a more direct measure of structural diversity, we performed a post-hoc analysis by computing the average internal Tanimoto similarity among the top-$100$ highest-reward molecules sampled from the final policies of each method. A lower Tanimoto similarity score indicates higher sample diversity.

The results are presented in Table~\ref{tab:tanimoto-results}. As this table shows, our algorithm does not sacrifice diversity while ensuring sampling optimality (achieving the highest Top-$100$ Reward). This strong balance of high reward and low internal similarity explains why, as shown in Figure 7, our method is able to discover a significantly larger number of distinct high-reward modes.

\begin{table}[h]
\caption{Comparison of Top-100 Reward (from Table 1) and Top-100 Internal Tanimoto Similarity (Avg. $\pm$ Std. over three seeds) on the Molecule Design task. Lower similarity indicates higher diversity.}
\label{tab:tanimoto-results}
\begin{center}
\begin{tabular}{lcc}
\toprule
\textbf{Method} & \textbf{Top-100 Internal Tanimoto Sim. ($\downarrow$)} & \textbf{Top-100 Reward ($\uparrow$)} \\
\midrule
Proxy-GFN & $0.665 \pm 0.024$ & $7.281 \pm 0.067$ \\
Oracle-GFN & $0.615 \pm 0.017$ & $\textbf{7.408} \pm \textbf{0.021}$ \\
\midrule[0.2pt]
Dataset-GFN & $0.521 \pm 0.026$ & $7.198 \pm 0.018$ \\
FM-COFlowNet & $0.535 \pm 0.015$ & $7.201 \pm 0.015$ \\
QM-COFlowNet & $0.526 \pm 0.014$ & $7.296 \pm 0.022$ \\
\textbf{TD-GFN (Ours)} & $0.531 \pm 0.022$ & $\textbf{7.450} \pm \textbf{0.037}$ \\
\bottomrule
\end{tabular}
\end{center}
\end{table}

\subsection{Naive Application of Conventional GFlowNet Objectives to Offline Datasets}
\label{apd:naive_offline}

To further investigate the necessity of our proposed framework, we evaluate the performance of standard GFlowNet training objectives---specifically Detailed Balance (DB)~\citep{bengio2023gflownet}, Trajectory Balance (TB)~\citep{malkin2022trajectory} and Sub-Trajectory Balance (SubTB, with $\lambda=0.9$)~\citep{madan2023learning}---when directly applied to the offline trajectory datasets without additional guidance. Experiments were conducted on the $8^4$ Hypergrid task using both the Expert and Bad datasets ($1,500$ trajectories each).

\begin{figure}[h]
    \centering
    \includegraphics[width=0.85\linewidth]{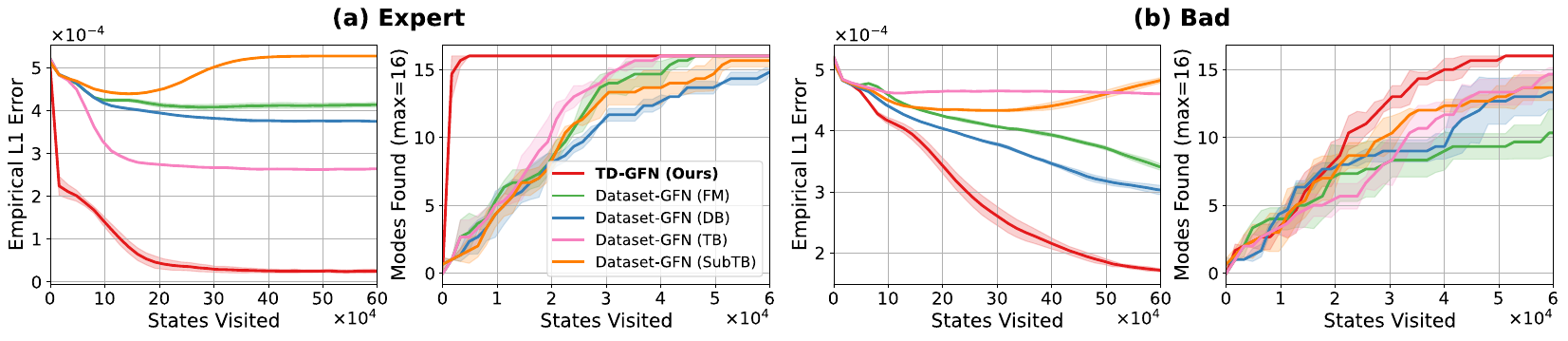}
    \caption{Performance comparison of conventional GFlowNet objectives when naively applied to $1,500$ offline trajectories in the \textit{Expert} and \textit{Bad} dataset on the $8^4$ Hypergrid task.}
    \label{fig:dataset_gfn}
\end{figure}

As illustrated in Figure~\ref{fig:dataset_gfn}, these conventional methods perform poorly compared to TD-GFN. In the absence of structural guidance for unobserved regions, simply restricting standard online objectives to fixed offline trajectories leads to suboptimal convergence and limited mode discovery. These results underscore that the naive transfer of online GFlowNet objectives to the offline setting is insufficient, highlighting the necessity of the specific mechanisms introduced in our framework.

\subsection{Learning curves in Biosequence Design task}
We provide the learning curves in Biosequence Design task in Figure~\ref{fig:bio_curve}. We omitted them in the main paper because, while TD-GFN remains stable throughout training, baseline methods exhibit significant volatility. To ensure a fair comparison, we reported the baseline results at the checkpoints that achieve the best balance between optimality and diversity.
\begin{figure}[htbp]
    \centering
    \includegraphics[width=0.5\linewidth]{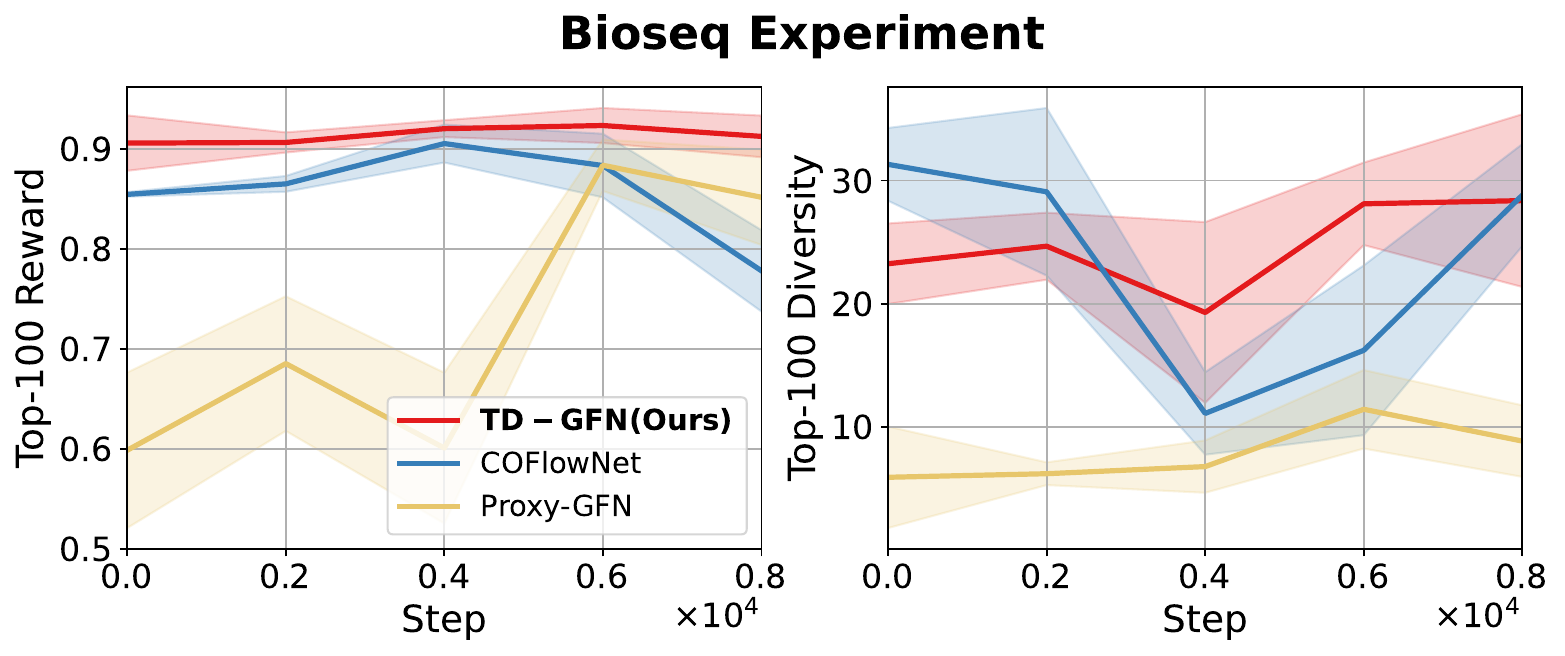}
    \caption{Learning curves of three methods in the Biosequence Design task.}
    \label{fig:bio_curve}
\end{figure}

\newpage
\section{Discussions}
\label{apd:discussion}

\paragraph{Adaptability to Extended GFlowNet Settings.} 
TD-GFN can be directly applied to scenarios involving non-acyclic directed graphs~\citep{brunswic2024theory}, as it operates over general transition structures without a fundamental reliance on acyclicity. In contrast, adapting our framework to continuous environments would necessitate additional modeling considerations (e.g., for defining and pruning over continuous edge sets). Furthermore, extending the method to handle stochastic rewards or non-deterministic transitions~\citep{pan2023stochastic} represents an important direction, particularly for real-world applications where uncertainty is inherent in both reward evaluation and environmental dynamics.

\paragraph{Toward More Advanced Edge Reward Modeling.} 
The current implementation of TD-GFN employs an adversarial IRL approach for edge reward estimation. This choice, while effective, 
inherits the well-known challenges of adversarial training, such as potential instability and sensitivity to hyperparameters. Therefore, the development of more stable IRL techniques presents a promising avenue for further enhancing the reliability and scalability of the TD-GFN framework. This may be achieved by leveraging recent advances in inverse reinforcement learning, such as stability-enhancing regularization~\citep{fu2018learning, Kostrikov2020Imitation}, representation learning–augmented IRL~\citep{Chandak2019LearningAR}, and score-based or energy-based IRL methods~\citep{Liu2020EnergyBasedIL}.

\paragraph{On Dataset Generation.}
Many of the datasets in our work are generated using GFlowNets. This strategy is also common in the offline reinforcement learning (RL) community~\citep{Fu2020D4RLDF,Gulcehre2020RLUB}, where datasets are typically collected by RL policies of varying quality. However, our method is not tied to any particular data generation procedure. For instance, in Section~\ref{sec:bioseq}, we employ a dataset curated from the DBAASP database~\citep{pirtskhalava2021dbaasp}. We adopt GFlowNet-generated datasets not because our approach requires them, but because they provide a controlled and reproducible way to simulate realistic scenarios in which data arises from behavior policies of different optimality levels. This allows for a systematic investigation of how dataset quality influences performance. Moreover, certain real-world datasets were not included simply because no proxy-based learning baselines are available for comparison on them. In future work, we aim to further validate the effectiveness of our approach on datasets that more closely resemble practical deployment environments.

\paragraph{What is Encoded in Trajectory Data.}
Our work is situated in the offline GFlowNet setting, where training is conducted on a fixed collection of reward-labeled trajectories. These trajectories—whether collected from humans, heuristics, or simulators—are often overlooked in proxy-based methods. In contrast, our approach is designed to leverage such information without resorting to proxy reward models.

In this setting, however, our method differs from proxy-based approaches in that it explicitly leverages the trajectories themselves and assumes that they encode valuable information beyond terminal rewards. A natural question, then, is what constitutes a \emph{valuable} trajectory. To deepen our understanding of the role of trajectories in offline GFlowNet training, we construct two additional datasets from the terminal states used in the main experiments by generating trajectories in reverse: (i) Uniform, where at each step a parent node is selected uniformly at random; and (ii) Rule-based, where the parent is chosen by preferentially reducing the dimension with the smallest current value, thereby producing trajectories that avoid the central regions of the hypergrid with low rewards. We then apply TD-GFN to these two datasets and compare the results with those obtained using the original GFlowNet-generated dataset. The results are presented in Figure~\ref{fig:datasets}.
\begin{figure}[htbp]
    \centering
    \includegraphics[width=0.85\linewidth]{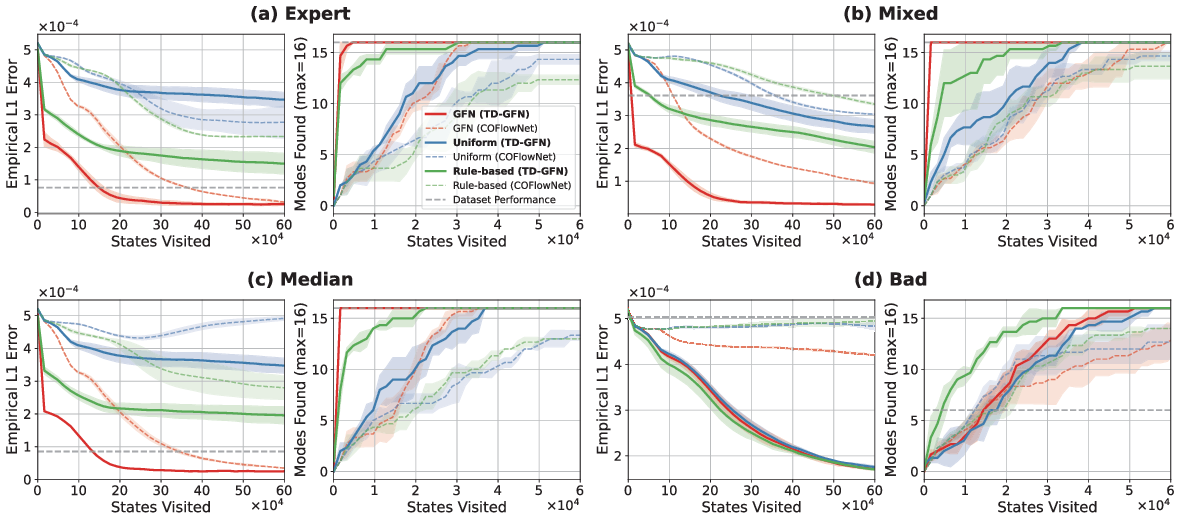}
    \caption{Performance comparison of TD-GFN policies trained on the $8^4$ Hypergrid task using three types of datasets with the same terminal states, each consisting of $1,500$ trajectories.}
    \label{fig:datasets}
\end{figure}

Although all three datasets share the same set of terminal states, the results clearly show that the trajectories themselves provide crucial information. Trajectories generated via uniform backward sampling perform very poorly, and even the rule-based trajectories, which are highly efficient and deliberately avoid low-reward regions, are less effective than trajectories generated forward by a GFlowNet. This observation suggests TD-GFN leverages meaningful trajectory structure, implying it may struggle with datasets consisting only of noise. Nevertheless, its sensitivity is mild compared to COFlowNet, which degrades more significantly under such conditions.

\paragraph{Future Directions}
The observation above opens up two promising directions for future work. First, developing principled methods for generating effective trajectories from terminal states could extend TD-GFN's applicability to settings where only terminal data is available.

Second, the strong performance of TD-GFN on GFlowNet-generated datasets strongly suggests its potential in online and interactive training settings. We envision that the edge rewards could be learned and updated dynamically as the agent collects new experience. This dynamic guidance model could then be used to continuously improve sampling efficiency for continued training, for instance, by intelligently focusing exploration on promising, high-reward regions of the DAG while steering the agent away from well-understood or low-utility pathways. 

Furthermore, an interesting future direction would be to replace the ``hard'' pruning of the DAG with a ``softer'' guidance mechanism. In fact, the pruning method can be formally understood as an extreme case of a ``soft'' guidance mechanism, such as re-weighting the policy logits using the learned edge rewards. In this view, pruning is equivalent to applying an infinitely large positive weight to transitions above the threshold and a zero weight (or infinitely negative logit) to those below it. This perspective provides a clear theoretical motivation, as this re-weighting can be interpreted as imposing a strong Bayesian prior on the policy (incorporating the structural knowledge learned via IRL) or as a form of imitation learning regularization that constrains the policy to align with the distilled expert behaviors. 

However, based on our initial explorations, while this ``soft'' method is a compelling alternative, it seems to be highly sensitive to its weighting-strength hyperparameter. A potential explanation is that it relies on the precise numerical accuracy of the learned edge rewards, which may contain generalization errors. Developing a more effective ``soft'' guidance mechanism remains a promising avenue for future research.

% \section{Statement on LLM Usage}
% \label{apd:llm_usage}
% In the spirit of transparency and in accordance with conference policy, we report our use of Large Language Models (LLMs) as assistive tools.

% We used LLM-based tools (e.g., GPT-4o and Gemini 2.5 pro) to polish the writing in this manuscript by improving grammar, clarity, and style. For software implementation, we used the AI-powered editor Cursor to assist with routine coding tasks, such as generating boilerplate code and auto-completion.

% In all instances, the LLMs served as productivity tools. The core research ideas, experimental design, model architecture, algorithmic logic, and interpretation of results are solely the work of the human authors, who bear full responsibility for all content in this paper.
\end{document}